\documentclass[lettersize,journal]{IEEEtran}
\usepackage{amsmath,amsfonts,amssymb}
\usepackage{algorithm}
\usepackage{algpseudocode}

\usepackage{graphicx}
\usepackage{hyperref}

\usepackage{array}
\usepackage{textcomp}
\usepackage[acronym]{glossaries}
\newacronym{cnn}{CNN}{convolutional neural network}
\usepackage{stfloats}
\usepackage{url}
\usepackage{verbatim}
\usepackage{graphicx}
\usepackage{cite}
\usepackage{bm}
\usepackage[capitalize]{cleveref}
\usepackage{xspace}
\usepackage[utf8]{inputenc}
\usepackage{booktabs}   

\usepackage{multirow}
\usepackage{makecell}
\usepackage{amsmath}    
\usepackage{graphicx}   

\usepackage{xcolor}

\newacronym{imu}{IMU}{inertial measurement unit}
\newacronym{tecdar}{TECDAR}{transient extrinsic contact detection and ranging}
\newacronym{ekf}{EKF}{extended kalman filter}
\newacronym{pla}{PLA}{polylactic acid}
\newacronym{pdms}{PDMS}{polydimethylsiloxane}
\newacronym{cad}{CAD}{computer aided design}
\newacronym{rmse}{RMSE}{root mean square error}
\newacronym{tcp}{TCP}{tool center point}
\newacronym{gd}{GD}{gradient descent}
\newacronym{snr}{SNR}{signal-to-noise ratio}
\newacronym{sd}{SD}{standard deviation}
\newacronym{svd}{SVD}{singular value decomposition}
\begin{document}

\title{Detection and Ranging of Transient Extrinsic Contacts Based on 6D Dynamic Tactile Sensing} 

\author{Haowen Zheng, Yinghao Wu, Fuyuan Liu, Yichen Li, and Yitian Shao
\thanks{This work was supported by National Natural Science Foundation of China under Grant 62576119 and Shenzhen Science and Technology Innovation Program under Grant JCYJ20241202123716021.}
\thanks{All authors are with the School of Computer Science and Technology, Harbin Institute of Technology, Shenzhen, Shenzhen 518055, China. \textit{(Corresponding author:
Yitian Shao.)}}}


\maketitle

\begin{abstract}
Delicate manipulation often involves transient and subtle collisions between a grasped object and the environment. 
While the human hand localizes these contacts effortlessly thanks to superior tactile sensitivity, robotic systems often lack the requisite resolution to acquire the information necessary for motion planning, resulting in clumsy manipulation or even task failure.
Here, we propose transient extrinsic contact detection and ranging (TECDAR), a simple yet fast and efficient method for detecting and ranging extrinsic contact of grasped objects. 
Our design of gripper tips employs dynamic tactile sensing leveraging a single 2.5$\times$3 mm 6D inertial measurement unit. The sensor captures sub-millisecond tip deformations at a 7 kHz sampling rate, but operating on a data stream of only 84\,KB/s. High bandwidth and compact data size enable the system to rapidly detect and localize contact between grasped objects and their surroundings.
Specifically, fusing tactile data with robot pose via an extended Kalman filter enables fast and precise localization of extrinsic contact, reaching millimeter-level accuracy within 180\,ms.
Experimental results demonstrate that the system achieves an average localization accuracy of approximately 7\,mm in both line-contact and point-contact localization tasks. Furthermore, this near-instantaneous localization enables the robot to rectify its trajectory on a millisecond scale, facilitating precise tool manipulation and enhanced perception of complex environments purely through tactile exploration and mapping.
We envision such techniques advancing the future of robotics across domains requiring delicate manipulation, including precision assembly, surgical assistance, and autonomous exploration in touch-dominant environments. Project page: \href{https://humitlab.github.io/TECDAR/}{humitlab.github.io/TECDAR/}
\end{abstract} 


\section{Introduction}
\IEEEPARstart{H}{uman} 
hand can naturally utilize a held tool as an extension of its haptic sensory processing. For instance, when placing a book onto a shelf that is too high to be visible, the hand can localize the boundaries of the shelf by decoding the subtle contact forces transmitted through the book to the fingertips. This tool-mediated tactile sensing transforms a passive object into an active sensory apparatus, enabling precise manipulation without visual feedback. Yet, replicating this human sensing capability in robotic systems remains underexplored.

While visual sensing plays a dominant role in enabling robotic systems to perceive the environment, it possesses inherent limitations in occluded manipulation tasks. Specifically, when a robotic arm holds an object and enters a confined space, the environment and the grasped object are often obscured from the visual field. This perceptual blind spot results in a loss of the critical contact state data required for motion planning, restricting the robot's ability to complete tasks in complex and unstructured environments.
Thereby, prior works have explored extrinsic contact localization using fingertip tactile sensing exclusively, successfully estimating the position of external line contacts on grasped objects \cite{bicchi2000,liang2024,ma2021extrinsic}. Specifically, when a grasped object interacts with a geometric constraint, such as a table edge, this physical interaction inevitably induces an instantaneous rotational tendency around the contact boundary. 
Existing methods predominantly focus on scenarios involving large rotation angles of the grasped object and slow fingertip deformation, which are typically captured using sensors with high spatial resolution, such as visuotactile sensors. Consequently, they require large data volumes and complex tracking algorithms, introducing a heavy computational workload.

In this work, we propose a responsive and data efficient framework, \gls{tecdar}, for localizing extrinsic contacts on grasped objects.
We adopt an event-driven sensing paradigm that captures both the instantaneous collision and the momentary rotations of a robotic fingertip upon extrinsic contact. 
Instead of capturing complex spatial pose information, our method relies on a single 6-axis \gls{imu} to perform high-temporal-resolution decoding of dynamic acceleration and angular velocity signals at a single sensing point on the robotic fingertip. Meanwhile, the miniaturized \gls{imu} ($2.5 \times 3 \times 0.8$\,mm) can be readily integrated into existing tactile devices without requiring extensive structural modifications.

Specifically, the accelerometer of the \gls{imu} provides low-latency on-set detection of contact events, acting as a precise trigger for the \gls{imu}'s gyroscope to capture transient rotational deformations of robot fingertip. This pipeline allows the robot to decode contact locations directly from the triggered angular velocity and gripper pose. Even under light contact resulting in minimal fingertip deformation, such a model-based approach enables efficient and swift state estimation, establishing a unified framework for real-time contact sensing with minimal computational overhead. 
This responsive localization capability enables robust, closed-loop control when interacting with objects of unknown rotational axes. We evaluate the control scheme by tasking a robotic manipulator with operating constrained mechanisms, such as a paper cutter, without prior geometric or kinematic knowledge. Real-time tracking of the rotation axis facilitates a closed-loop operation that dynamically corrects the end-effector trajectory, thereby preventing excessive joint stress or structural damage to the manipulated target. 
Moreover, using this contact localization capability allows the robot to probe an unknown region through repeated contact, incrementally forming a ``haptic map'' of the environment. This tactile exploration helps the robot navigate complex, unstructured environments even when visual sensing is obscured or occluded.


\begin{figure*}[t]
    \centering
    \includegraphics[width=178mm]{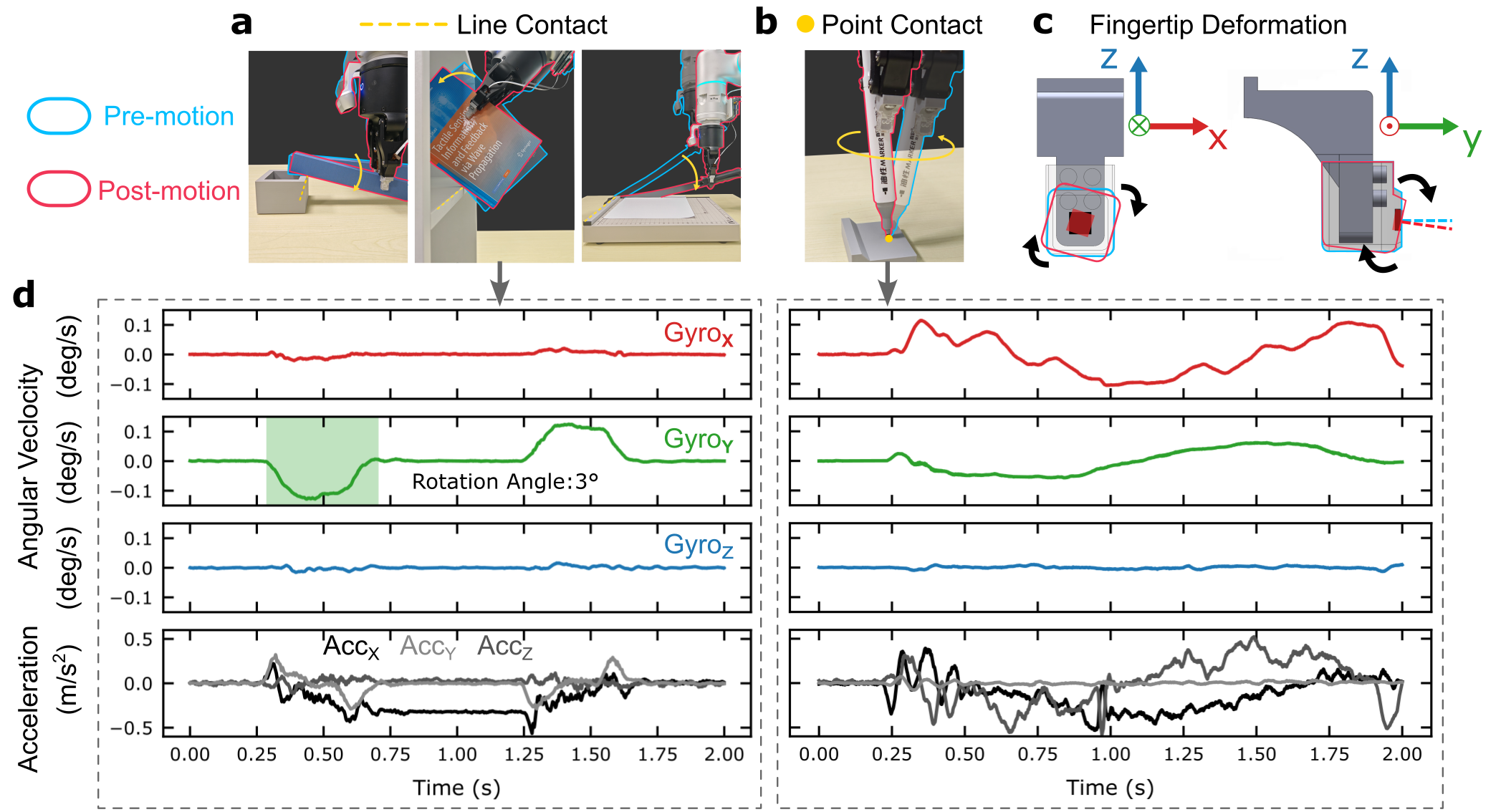} 
    \caption{\textit{Tactile sensing framework and representative interaction cases.} The top panel illustrates four experimental scenarios with pre- and post-movement states: (a) a plastic prism for controlled baseline validation, a book simulating real-world line contact, and a paper cutter demonstrating constrained trajectory planning; and (b) a rotating pen representing a typical point-contact case. (c) shows the instantaneous rotational states, and (d) the corresponding gyroscope signal responses for the cases of the rotating book and pen. The embedded gyroscopes in the robot fingertip can capture high-frequency angular rate pulses (e.g., transients in $Gyro_y$), which serve as the key tactile cues to estimate the spatial coordinates of the contact line or point.}
    \label{fig:concept}
\end{figure*}

The main contributions of this paper are summarized as follows:
\begin{itemize}
    \item A novel tactile sensing mechanism featuring low data rates, low cost, and compact form factor: We present a tactile sensing approach utilizing only a single 6-axis \gls{imu} that bypasses the need for high-data volume visuo-tactile or expensive force/torque sensors
    We introduce a minimalist tactile sensing framework leveraging a single 6-axis \gls{imu} to circumvent the need for data-heavy visuo-tactile sensors and expensive force/torque sensors, achieving robust perception with significantly-reduced computational overhead.
    \item Fast and efficient event-driven extrinsic contact sensing framework purely based on fingertip deformation: We establish a synergistic pipeline where the accelerometer isolates the precise moment of collision, and a differential kinematic model translates high-frequency fingertip torsional transients into spatial coordinates, enabling the rapid localization of extrinsic contact points during the initial moments of interaction.
    \item Real-time trajectory control with latency as low as 20 ms: We demonstrate that the proposed sensing mechanism allows the robotic system to perform on-the-fly trajectory correction at a 20 ms control response cycle. This enables the smooth manipulation of a constrained mechanism with an unknown rotation axis, ensuring stable interaction through immediate kinematic feedback.
    \item An efficient, Bayesian-informed geometric mapping framework that relies exclusively on touch interaction: By generating probabilistic geometric representations through pure tactile exploration, our framework enables autonomous robotic navigation and manipulation within complex, unstructured environments in the absence of vision.
\end{itemize}

\section{Related Work}

\subsection{Tactile Sensing for Robotic Manipulation}

Force/torque sensors are well-established in industrial applications \cite{cao2021six}, providing global force information, yet the large form factor limits their integration into the robot fingertips. 
Piezoresistive, piezoelectric, and capacitive tactile sensors feature low-profile designs well-suited for integration into robotic skin or fingertips \cite{luo2021learning, bhirangi2025anyskin}, providing a versatile hardware substrate for distributed tactile sensing. Yet, their complex material structures and electronics often pose significant challenges in maintaining sensing consistency, long-term durability, cost efficiency, and mass production \cite{huo2026recent}.
Visuotactile sensors have been widely adopted due to their high spatial resolution, ease of fabrication, and excellent repeatability \cite{yuan2017gelsight, lambeta2020digit}, leveraging embedded cameras to capture soft elastomer deformations for the precise perception of contact locations, surface textures, and force distributions. However, these sensors prioritize spatial information over temporal information, and variants that adapt event cameras to enlarge sensing bandwidth face challenges of high cost and bulky size \cite{funk2024evetac, yin2025gelevent}.
Acoustic sensors, such as piezoelectric contact microphones, complement tactile perception from the temporal dimension \cite{taunyazov2021extended, schurmann2012high}, with their high-frequency response enabling the capture of transient vibration signals at the moment of contact, proving particularly effective for slip detection and contact event recognition. Recent work demonstrated the utility of fingertip-mounted \gls{imu} signals for slip detection in compliant robotic hands, highlighting the potential of low-cost inertial transient sensing \cite{cravetz2025slip}.

\subsection{Extrinsic Contact Localization}

In robotic manipulation tasks, accurate estimation of contact point locations when a grasped object interacts with the environment provides critical state information for subsequent motion planning and force control. Kinematic constraint optimization formulates contact estimation by recovering contact geometry from relative motion tracked by distributed tactile measurements \cite{ma2021extrinsic}. Factor-graph formulations extend this model-based approach, fusing kinematic data with tactile measurements and introducing active control strategies that improve estimation accuracy in tasks such as peg-in-hole insertion \cite{kim2022ActiveExtrinsicContact}. These model-based approaches offer strong interpretability and low dependence on training data. 

Data-driven methods pursue an alternative path, leveraging high-dimensional tactile measurements and learned representations. Implicit neural representations map sequences of tactile measurements to contact probability distributions over three-dimensional object surfaces, achieving extrinsic contact tracking for arbitrarily shaped objects with promising sim-to-real transfer \cite{higuera2023NeuralContactFields}. Geometry-agnostic methods estimate the direction vector of extrinsic line contacts by tracking relative object motion through high-resolution tactile images \cite{kimExtrinsicLineContact2025}. Simultaneous estimation frameworks address contact location and object pose jointly using proprioception and tactile feedback \cite{sipos2022SimultaneousContactLocation, kim2023SimultaneousTactileEstimation}. Single-image methods infer six-degree-of-freedom object pose from a single tactile image, achieving sub-millimeter accuracy in specific tasks \cite{bauza2023Tac2PoseTactileObjecta}, while cross-sectional reconstruction approaches achieve zero-shot pose estimation through point cloud registration with pre-stored CAD models \cite{yajima2026Touch2InsertZeroShotPeg}. Complementary efforts extend to contact patch estimation for stable placement \cite{ota2024TactileEstimationExtrinsic} and visuo-tactile implicit representations for simultaneous in-hand pose and extrinsic contact estimation \cite{leeViTaSCOPEVisuotactileImplicit2025}.

However, most existing works rely on dense sensor arrays or high-dimensional visual processing. Exploiting low-cost inertial sensing to capture the instantaneous rotational signals induced by extrinsic contact at the fingertip remains unexplored.


\subsection{Tactile-Driven Control and Planning}

In the field of trajectory planning and control under contact constraints, two distinct technical routes have emerged. Contact-implicit trajectory optimization embeds contact dynamics directly into the optimization framework, allowing the solver to automatically discover contact making and breaking without pre-specified contact mode sequences \cite{posa2014direct}. Variational integrator formulations extend this approach with improved numerical stability and energy conservation for long-horizon planning \cite{manchester2019variational}. From a geometric perspective, motion cone methods explicitly parameterize feasible motion directions under contact constraints as convex cones, enabling planners to rapidly identify physically admissible motions given known contact configurations \cite{chavan2020planar}. While these optimization-based and geometry-based routes differ in formulation, they share a common characteristic: both handle contact constraints at the planning level and typically require offline computation or prior knowledge of contact configurations.

In the domain of contact force control and physical interaction, the Series Elastic End Effector (SEED) model captures the compliance of high-resolution tactile sensors as linear elastic elements, establishing an approximate linear relationship between sensor deformation and transmitted force/torque \cite{suh2022seed}. This transforms contact tasks that originally required precise force control into position control problems, substantially simplifying force-interaction control. Differentiable quadratic programming frameworks further extend this approach by combining the tactile elasticity model with friction cones and force balance, achieving non-prehensile manipulation skills such as pushing and pivoting with a grasped tool \cite{ollerTactileDrivenNonPrehensileObject2024}. Factor-graph-based contact configuration regulation fuses multi-modal sensor data to estimate contact states in real time and adjust control actions accordingly, enabling stable manipulation under multiple and intermittent contacts \cite{taylorObjectManipulationContact2023}. These works share a reliance on tactile feedback as a core control-loop input, transforming perceptual information into control actions through physical modeling or probabilistic inference.

For robotic touch powered by deep learning, tactile-RL achieves insertion that generalizes to objects of unknown geometry \cite{dong2021tactile}, and learned policies extend to tactile tool manipulation \cite{shiraiTactileToolManipulation2023}. Recent efforts advance dexterous in-hand tasks, including simultaneous tactile estimation and control for extrinsic dexterity \cite{bronars2024TEXterityTactileExtrinsic}. Taxel-array-based sensing enables dexterous in-hand writing via extrinsic contact perception \cite{zhaoTactileDrivenDexterousInHand2025}, while contact-deviation reframing treats robot-object interactions as rich sources of local kinematics for proactive articulated object manipulation \cite{zhao2026tacman}. Visuo-tactile implicit representations handle complex geometries through simultaneous in-hand pose and extrinsic contact estimation \cite{leeViTaSCOPEVisuotactileImplicit2025}. These works demonstrate tactile perception extending beyond localization into closed-loop control.
In contrast to the aforementioned approaches, the framework proposed in this work prioritizes millisecond-level sensing and response.
Rather than relying on offline trajectory optimization or learned policies, we recursively minimize the differential constraint objective at each time step through Bayesian filtering, directly coupling tactile perception with real-time control. This allows the proposed method to serve as a low-level feedback module that provides real-time geometric constraint information for contact-rich tasks.

\section{Method} 

Our approach is driven by three key objectives: First, to capture the subtle rotations occurring at the fingertip during extrinsic contact through a low-cost, compact gyroscopic sensor. Second, to utilize a physically-constrained model that maps these rotations and the gripper’s pose to precise contact locations via differential kinematic transforms. Third, to ensure high temporal resolution and low latency, permitting the detection of extrinsic contacts even during short-duration, small-scale rotary deformations. By reformulating tactile perception as a low-dimensional state estimation task, we achieve a sensing mechanism that is both mechanically integrated and computationally streamlined.


\subsection{Hardware Design and Integration}
To achieve high-frequency transient event detection in a minimal form factor, we build upon the TranTac vibration-sensing principle~\cite{wu2025TranTacLeveragingTransient} and present a substantially miniaturized iteration of its core design. Crucially, our sensor is not intended to replace existing high-resolution modalities but to complement them. Its ultra-low data throughput and high temporal resolution make it an ideal companion sensor: while visuotactile sensors provide steady-state contact geometry, the proposed system captures high-frequency transient events at the moment of contact. The two modalities work together to enhance the robustness and responsiveness of robotic perception in contact-rich scenarios. Table~\ref{tab:tactile_comparison} quantitatively contextualizes the proposed sensor within the landscape of existing tactile sensing systems that has been adopted for extrinsic contact localization.
Vision-based tactile sensors such as GelSlim~3.0 achieve sub-millimeter localization accuracy, but this precision relies on high-resolution imaging that generates over 27\,MB/s of data, restricting their deployment in time-critical control loops. 
\gls{tecdar} inverts this trade-off: its localization error is higher (5.36\,mm), but with sensor data volume dropped by more than 300-fold to 84\,KB/s, with temporal resolution reaching 0.14\,ms.
The localization recursion operates on six-dimensional state vectors, adding negligible computational and telemetry overhead to the control loop.

\begin{table}[tb]
    \centering
    \caption{\textnormal{Comparison of the \gls{tecdar} with state-of-the-art tactile sensing systems for extrinsic contact localization. Performance indicators show whether $\downarrow$ low values are better or $\uparrow$ high values are better. The best values are in \textbf{bold}.}}
    \label{tab:tactile_comparison}
    \begin{tabular}{>{\raggedleft\arraybackslash}p{18mm}p{12mm}p{12mm}p{12.5mm}p{12mm}}
        \toprule
        Sensor/System & \gls{tecdar} (ours) & GelSlim 3.0 \cite{kim2023SimultaneousTactileEstimation} & 
GelSlim 3.0 \cite{ma2021extrinsic} & GelSight Mini \cite{van2026simultaneous} \\
        \midrule
$\downarrow$ \hfill Size [mm] & \textbf{15$\times$10$\times$12} & 37$\times$80$\times$20 & 37$\times$80$\times$20 & 30$\times$25$\times$20 \\
\midrule
$\downarrow$ \hfill Cost [\$] & \textbf{5} & 25 & 25 & 10 \\
\midrule
$\downarrow$ \hfill Axes/pixels & \textbf{6} & 640$\times$480 & 640$\times$480 & 640$\times$480 \\
\midrule
$\uparrow$ \hfill Bandwidth [Hz] & \textbf{7000} & 30 & 30 & 30 \\
\midrule
$\downarrow$ \hfill Data Volume [KB/s] & \textbf{84} & 27648 & 27648 &  27648 \\
\midrule
$\downarrow$ \hfill Localization Error [mm] & 5.36$\pm$1.91 & 0.89$\pm$0.4 & \textbf{0.45$\pm$0.21}$^{\ast}$ & 3.72$\pm$3.78 \\
\midrule
Model & Bayesian Filtering & Deep Learning & Least Squares & Deep Learning \\
        \bottomrule
\multicolumn{5}{p{\dimexpr\columnwidth-2\tabcolsep}}{$^{\ast}$ result recreated from a single plot.} \\
    \end{tabular}
\end{table}

\begin{figure}[t]
    \centering
    \includegraphics[width=1\linewidth]{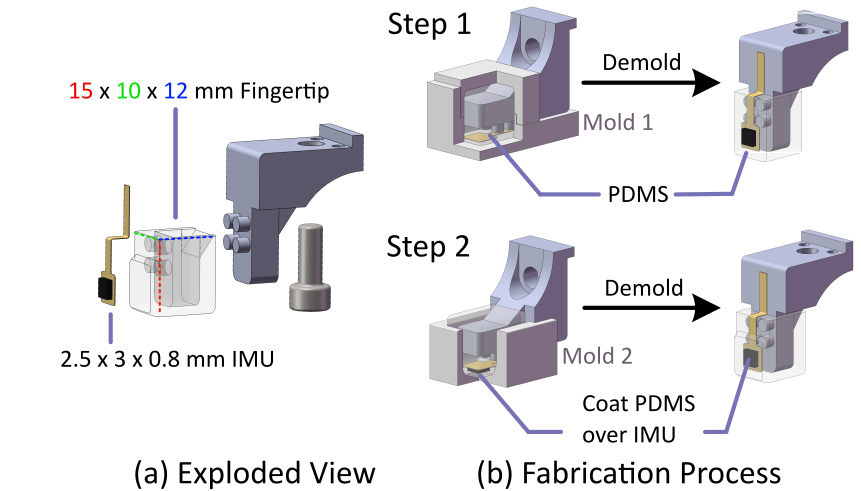}
    \caption{Hardware design and manufacturing of the \gls{tecdar} tactile sensor. (a) Exploded view illustrating the compact integration of the miniature \gls{imu} chip ($2.5 \times 3 \times 0.8$~mm) at the geometric center of the compliant PDMS fingertip. (b) The two-step molding process ensuring complete encapsulation of the sensor for stable strain transmission.}
    \label{fig:fabrication}
\end{figure}

We introduce a simple yet effective sensing approach for detecting transient fingertip deformations of a robot gripper, leveraging the high sensitivity of accelerometers and gyroscopes to capture the subtle tactile signals produced by translational and rotational motion of the elastomeric skin.
Specifically, our system adopts a six-axis inertial sensor (ST LSM6DSR iNEMO) with a highly compact package size of $2.5 \times 3 \times 0.8$~mm as the core sensing unit. As illustrated in the exploded view in \cref{fig:fabrication}(a), this miniature \gls{imu} is integrated into a $15 \times 10 \times 12$\,mm fingertip structure. {Our framework leverages the complementary modalities of the 6-axis \gls{imu}: the accelerometer channel captures high-frequency impact vibrations for instantaneous contact on-set detection, and the gyroscope channel performs spatial kinematic estimation.} This unit offers a full-scale range of \(\pm 4000\),dps and a sampling rate of approximately $7$\,kHz, enabling accurate capture of angular transients induced by fingertip deformation.

The structural design and fabrication process are illustrated in \cref{fig:fabrication}. The \gls{imu} is embedded at the center of the fingertip’s elastomeric contact region, as shown in \cref{fig:fabrication}(b); its placement is consistent with the contact area slip distribution analysis by Li et al.~\cite{li2024IncipientSlipBasedRotation}. This structural configuration, combined with the tailored fingertip geometry, magnifies dynamic bending, shear, and torsional responses upon contact, facilitating signal acquisition by the sensor. This design enables accurate estimation of the kinematics of objects grasped by the fingertip.


Specifically, extrinsic contact generates on the fingertip a contact area composed of a peripheral slip zone surrounding a central stick region. Within this stick region, the no-slip constraint ensures that the local material deformation remains coupled with the rigid-body motion of the grasped object, making it the optimal site for tracking the motion.
Centering the sensor minimizes slip-induced noise during full contact with the elastomeric fingertip, assuming that most everyday objects are large enough to engage the entire
10$\times$10\,mm fingertip area. This allows the IMU to precisely capture the object's rotational motion within the gripper's coordinate system.

The sensor is encapsulated within a \gls{pdms} medium via a two-step molding process, as detailed in \cref{fig:fabrication}(b). This approach adopts the embedding strategy introduced in the TranTac system~\cite{wu2025TranTacLeveragingTransient}.
Specifically, the main PDMS body is first cast using Mold 1 (Step 1). Subsequently, the sensor flex board is positioned, and a second casting using Mold 2 coats the sensor (Step 2), ensuring complete encapsulation. 
The \gls{pdms} functions as both a protective layer and a continuous strain-transmission medium. Localized constraints from external contact at the fingertip surface induce a spatially distributed shear strain field; through material coupling, this field drives the embedded chip to undergo rotations, which the gyroscope then converts into angular velocity data.
Concurrently, the damping characteristics of \gls{pdms} physically suppress local high-frequency perturbations, making the observed signal better approximate the global rigid-body rotational response rather than localized deformation noise, thereby enhancing consistency between the observation model and the kinematic assumptions.

The proposed design detects transient fingertip rotations resulting from sudden extrinsic contact of a grasped object. The \gls{imu}’s accelerometer detects the transient spike produced by physical impact; this acts as a trigger, activating the localization module to process the gyroscope data that tracks the changing angular velocity of the fingertip with ultra-low latency.
The angular velocity is synchronized with the robot's proprioceptive state with a sampling frequency of 65\,Hz, ensuring precise temporal consistency in state estimation. In the subsequent modeling phase, the angular velocity is functionally related to the contact point position through rigid-body kinematic constraints, enabling recursive estimation of the contact location.


\subsection{Kinematic Modeling}

\begin{figure}[t]
    \centering
    \includegraphics[width=68mm]{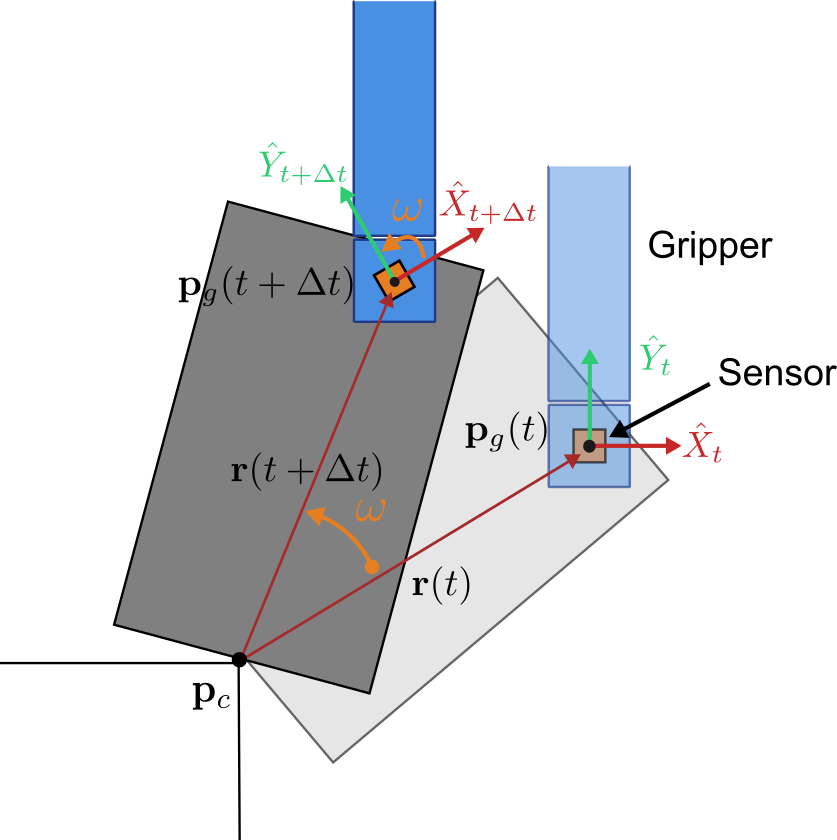}
    \caption{Kinematic modeling of extrinsic contact localization based on differential constraints. Under the rigid-body assumption, the instantaneous linear velocity $\mathbf{v}_{g}(t)$ of the tool center point $\mathbf{p}_{g}(t)$ is geometrically coupled with the angular velocity $\boldsymbol{\omega}(t)$ and the relative position $\mathbf{p}_{g}(t) - \mathbf{p}_c$. By capturing these motion transients, a differential kinematic mapping is established to isolate and estimate the unknown contact point $\mathbf{p}_c$ using the known robot kinematics and sensor measurements.}
    \label{fig:modeling}
\end{figure}

The core of extrinsic contact localization lies in identifying the kinematic constraints imposed on the grasped object in space. An analytical model can be constructed by assuming a rigid-body relationship among the grasped object, the gripper, and the environmental contact point.

\subsubsection{Geometric Constraints and Optimization}
When an object maintains a fixed point contact with the environment without sliding, it rotates around this environmental contact point. Under this condition, the instantaneous center of rotation coincides with the contact point $\mathbf{p}_c$. Let $\mathbf{p}_c = [x_c, y_c, z_c]^T \in \mathbb{R}^3$ denote the static contact position vector to be estimated. Let the position vector of the \gls{tcp} at time $t$ be $\mathbf{p}_{g}(t) = [x_g(t), y_g(t), z_g(t)]^T \in \mathbb{R}^3$. We define the relative position vector of the \gls{tcp} with respect to the contact point as $\mathbf{r}(t) = \mathbf{p}_g(t) - \mathbf{p}_c$. The magnitude of this vector, $L = \|\mathbf{r}(t)\|_2$, represents the rotation radius, which dictates the kinematic scaling of the interaction. During this process, the incremental rotation of the object around $\mathbf{p}_c$ is described by the matrix $\mathbf{R}(\Delta t) \in SO(3)$. Based on geometric consistency, the spatial vector of the \gls{tcp} relative to the rotation center must satisfy the following displacement constraint equation:
\begin{equation}
    \mathbf{r}(t+\Delta t) = \mathbf{R}(\Delta t) \mathbf{r}(t). \label{eq:geo_consistency}
\end{equation}
Based on the geometric constraints above, the contact point estimation can be formulated as an optimization problem. Prior work localized extrinsic contact by formulating it as a least-squares fitting problem and solved using \gls{svd} \cite{ma2021extrinsic}. 
Specifically, by minimizing the cumulative geometric residual over the observation period, the contact point $\mathbf{p}_c$ is estimated as follows:
\begin{equation}
    \mathcal{L}_{\mathrm{geo}}(\mathbf{p}_c) = \frac{1}{2} \int \left\| (\mathbf{p}_{g}(t) - \mathbf{p}_c) - \mathbf{R}(t)(\mathbf{p}_{g}(0) - \mathbf{p}_c) \right\|_2^2 dt, \label{eq:geo_loss}
\end{equation}
where $\mathbf{R}(t)$ is the absolute rotation matrix integrated from $t=0$. This rotation matrix cannot be directly measured and must be obtained by integrating the angular velocity signals output by the gyroscope. 
However, the integration process inevitably accumulates sensor noise and drift, making the direct optimization method based on geometric constraints sensitive to measurement errors. Therefore, this geometric method serves primarily as a baseline for subsequent experimental comparisons.

\subsubsection{Differential Constraints and Instantaneous Kinematics}

To ensure more robust state estimation, we propose a differential form of kinematic constraints. This formulation circumvents the time-integration of angular velocity, thereby isolating the estimation from the cumulative drift inherent in gyroscopic bias. We expand and rearrange \cref{eq:geo_consistency} into the form of displacement increments:
\begin{equation}
    \mathbf{r}(t+\Delta t) - \mathbf{r}(t) = (\mathbf{R}(\Delta t) - \mathbf{I}) \mathbf{r}(t), \label{eq:disp_inc}
\end{equation}
where $\mathbf{I} \in \mathbb{R}^{3 \times 3}$ is the identity matrix. This equation geometrically describes the kinematic constraint of the end-effector trajectory: the displacement increment of the \gls{tcp} is restricted within a spatial envelope centered at $\mathbf{p}_c$.

The gyroscope integrated in each fingertip can measure the instantaneous angular velocity of fingertip deformation $\boldsymbol{\omega}(t) = [\omega_x(t), \omega_y(t), \omega_z(t)]^T \in \mathbb{R}^3$. This variable characterizes the deviation from unconstrained motion when the grasped object is subjected to the kinematic constraints of extrinsic contact. As illustrated in \cref{fig:modeling}, under the rigid-body assumption, the angular velocity of the \gls{tcp} corresponds to the rotation of the grasped object about the pivot point $\mathbf{p}_c$. 

For an infinitesimally small sampling period $\Delta t \to 0$, the incremental rotation matrix $\mathbf{R}(\Delta t)$ can be linearized via the first-order Taylor expansion of the exponential map:
\begin{equation}
    \mathbf{R}(\Delta t) \approx \mathbf{I} + [\boldsymbol{\omega}(t) \Delta t]_{\times}. \label{eq:exp_map}
\end{equation}

Substituting \cref{eq:exp_map} into the displacement increment equation \cref{eq:disp_inc} yields:
\begin{equation}
    \mathbf{r}(t+\Delta t) - \mathbf{r}(t) \approx [\boldsymbol{\omega}(t) \Delta t]_{\times} \mathbf{r}(t). \label{eq:diff_sub}
\end{equation}

Dividing both sides by $\Delta t$ and taking the limit as $\Delta t \to 0$, the left-hand side defines the instantaneous linear velocity $\mathbf{v}_g(t) \in \mathbb{R}^3$ of the \gls{tcp}. The differential constraint equation describing the instantaneous velocity field is:
\begin{equation}
    \mathbf{v}_{g}(t) = \boldsymbol{\omega}(t) \times \mathbf{r}(t). \label{eq:vel_field}
\end{equation}

\cref{eq:vel_field} establishes a direct mapping between the observed motion and the unknown contact point $\mathbf{p}_c$. Defining the instantaneous velocity residual, the contact point estimation problem is transformed into minimizing the differential kinematic loss:
\begin{equation}
    \mathcal{L}_{\mathrm{diff}}(\mathbf{p}_c) = \frac{1}{2} \int \left\| \mathbf{v}_{g}(t) - \boldsymbol{\omega}(t) \times (\mathbf{p}_{g}(t) - \mathbf{p}_c) \right\|_2^2 dt. \label{eq:diff_loss}
\end{equation}

\subsection{State Estimation and Filtering}

While the differential constraint optimization objective $\mathcal{L}_{\mathrm{diff}}(\mathbf{p}_c)$ can theoretically be solved via gradient descent, its inherent sensitivity to noise limits its practical utility for real-time applications. 
Leveraging the Markov property of the kinematic system and targeting a lightweight algorithm for real-time localization, we formulate the contact point estimation problem based on \gls{ekf}. 
For real-time implementation on digital systems, the continuous-time kinematic models derived in the previous section are discretized, where $k$ denotes the discrete sampling step. To prevent the accumulation of gyroscope drift during non-contact phases, we adopt an event-driven approach: the \gls{ekf} is initialized ($k=0$) only when the accelerometer measurement exceeds a predefined threshold, signaling contact onset. Extrinsic contact generates a physical impulse that enables our detection method to trigger events with ultra-low latency, typically on the order of 15\,ms.

Once triggered, the differential constraint \cref{eq:vel_field} is incorporated as the observation model, allowing for the recursive minimization of velocity residuals within a Bayesian framework. The corresponding true augmented state vector at discrete time step $k$ is defined as
\begin{equation}
    \mathbf{x}[k] = [\mathbf{p}_g[k]^T, \mathbf{v}_g[k]^T, \mathbf{p}_c[k]^T]^T \in \mathbb{R}^9, \label{eq:state_vector}
\end{equation}
where $\mathbf{p}_g[k]$ and $\mathbf{v}_g[k]$ represent the position and linear velocity of the \gls{tcp} in the world frame, respectively, and $\mathbf{p}_c[k]$ represents the coordinates of the extrinsic contact point to be estimated. Following the derivation in the previous section, the discrete-time relative pose vector is denoted as $\mathbf{r}[k] = \mathbf{p}_g[k] - \mathbf{p}_c[k]$.

$\mathbf{p}_g[k]$ and $\mathbf{v}_g[k]$ are derived directly from the robot's proprioceptive forward kinematics rather than through the integration of the \gls{imu}'s accelerometer. While the accelerometer is highly effective for detecting impact transients for event triggering, utilizing its signal for low-speed kinematic tracking leads to significant quadratic drift. This is caused by the inherent bias instability and low-frequency noise that inevitably accumulate during the integration of acceleration signals. In contrast, the robot's high-resolution joint encoders provide precise \gls{tcp} state information. Thus, our design implements a complementary sensing strategy that maximizes the reliability and robustness of the observation model. Robot proprioception provides low-drift motion tracking, while the \gls{imu} captures the micron-level transient dynamics of the fingertip deformation by utilizing the accelerometer for impact triggering and the gyroscope for rotational tracking. 
In practice, because the robot's proprioceptive sampling rate (around 65 Hz) is far lower than the EKF update rate (7 kHz), $\mathbf{p}_g[k]$ and $\mathbf{v}_g[k]$ are linearly interpolated between consecutive measurements to ensure continuous kinematic input at every time step.

The prediction stage follows a discretized transition equation based on kinematic models:
\begin{equation}
    \hat{\mathbf{x}}^{-}[k] = \mathbf{\Phi} \hat{\mathbf{x}}^{+}[k-1], \label{eq:prediction}
\end{equation}
where $\hat{\mathbf{x}}^{-}[k]$ denotes the a priori state estimate. During the prediction step, the state transition matrix $\mathbf{\Phi}$ integrates the kinematics over the sampling period $\Delta t$. 
To accurately reflect the rigid body motion, the \gls{tcp} velocity vector is explicitly rotated by the angular displacement $\Delta \theta = \omega_y[k] \Delta t$ measured by the gyroscope. The transition matrix is formulated as the following block matrix:
\begin{equation}
    \mathbf{\Phi} = \begin{bmatrix} 
    \mathbf{I} & \Delta t \mathbf{I} & \mathbf{0} \\ 
    \mathbf{0} & \mathbf{R}(\Delta \theta) & \mathbf{0} \\ 
    \mathbf{0} & \mathbf{0} & \mathbf{I} 
    \end{bmatrix}, \label{eq:transition_matrix}
\end{equation}
where $\mathbf{I}$ and $\mathbf{0}$ represent the identity and zero matrices of appropriate dimensions, respectively, and $\mathbf{R}(\Delta \theta)$ denotes the standard rotation matrix corresponding to the angular displacement.
Alongside the state prediction, the error covariance matrix $\mathbf{P}$ is propagated as:

\begin{equation}
    \mathbf{P}^{-}[k] = \mathbf{\Phi} \mathbf{P}^{+}[k-1] \mathbf{\Phi}^T + \mathbf{Q}, \label{eq:cov_predict}
\end{equation}
where the process noise covariance matrix $\mathbf{Q} = \text{diag}(\mathbf{Q}_p, \mathbf{Q}_v, \mathbf{Q}_c)$ characterizes the uncertainty of the system model. Specifically, $\mathbf{Q}_p$ and $\mathbf{Q}_v$ reflect random perturbations at the actuator level.

Although the contact point $\mathbf{p}_c$ is theoretically static, we explicitly set its process noise covariance $\mathbf{Q}_c > \mathbf{0}$. During the initial contact phase, low signal-to-noise ratios or insufficient kinematic excitation may lead to an inaccurate early estimate. If $\mathbf{Q}_c$ were strictly zero, the filter would quickly lock onto this incorrect initial value and reject subsequent updates. Introducing a small process noise ensures the filter remains responsive, continuously refining the estimate as more reliable sensor data arrives. 
To prevent filter divergence, a minor process noise component is introduced even for a nominally static state \cite{bar2001estimation}.

The observation update stage calibrates the state through a nonlinear mapping $h(\cdot)$ of multi-source sensor information. The observation vector $\mathbf{z}[k]$ includes the proprioceptive \gls{tcp} position $\mathbf{p}_g[k]$ and the instantaneous linear velocity $\mathbf{v}_g[k]$ based on differential constraints. The observations are provided as:
\begin{equation}
    \mathbf{z}[k] = h(\mathbf{x}[k]) = \begin{bmatrix} \mathbf{p}_g[k] \\ \mathbf{v}_g[k] \end{bmatrix}, \label{eq:obs_operator}
\end{equation}
where the velocity observation $\mathbf{v}_g[k]$ originates from the aforementioned differential kinematic equation \cref{eq:vel_field}. Since we assume $\mathbf{p}_c$ is the instantaneous rotation center, the linear velocity of the \gls{tcp} is entirely determined by the cross product of the effective angular velocity $\boldsymbol{\omega}[k]$ and the relative pose vector $\mathbf{r}[k]$. 

Due to the coupling between the state variables $\mathbf{p}_g[k]$ and $\mathbf{p}_c[k]$ in the observation equation, linearization must be performed via Taylor series expansion. At each sampling moment, the Jacobian matrix $\mathbf{H}[k]$ of the observation function with respect to the state vector is calculated:
\begin{equation}
    \mathbf{H}[k] = \left. \frac{\partial h(\mathbf{x})}{\partial \mathbf{x}} \right|_{\hat{\mathbf{x}}^{-}[k]}  = \begin{bmatrix} \mathbf{I}_{3 \times 3} & \mathbf{0}_{3 \times 3} & \mathbf{0}_{3 \times 3}\\ [\boldsymbol{\omega}[k]]_{\times} & \mathbf{0}_{3 \times 3} & -[\boldsymbol{\omega}[k]]_{\times} \end{bmatrix},
    \label{eq:jacobian_def}
\end{equation}
where $[\boldsymbol{\omega}[k]]_{\times}$ is the skew-symmetric matrix corresponding to the effective angular velocity vector. This matrix transforms the 3D cross-product operation into matrix multiplication, representing the contribution rate of the contact point deviation to the velocity residual within the linearized framework.

The corresponding measurement noise covariance $\mathbf{M}$ describes the random error statistics of the sensors. The position observation component is set based on the variance of the robot end-effector's repeatability, while the velocity component includes the equivalent variance of the gyroscope's white noise propagated through the nonlinear operator.

After completing the observation linearization based on the current predicted value $\hat{\mathbf{x}}^{-}[k]$, the system enters the stage of recursive update to achieve optimal correction of the state prediction by minimizing the posterior estimation error covariance, which is mathematically equivalent to solving the differential constraint optimization objective. The measurement innovation $\mathbf{y}[k]$ is computed as:
\begin{equation}
    \mathbf{y}[k] = \mathbf{z}[k] - h(\hat{\mathbf{x}}^{-}[k]), \label{eq:innovation}
\end{equation}
here, $\mathbf{y}[k]$ captures the kinematic mismatch during extrinsic contact.
If the linear velocity component in the residual is non-zero, it indicates a geometric conflict between the estimated contact point $\hat{\mathbf{p}}_c[k]$ and the instantaneous velocity field generated by $\boldsymbol{\omega}[k]$. This conflict is precisely the non-zero manifestation of the residual term in the optimization objective $\mathcal{L}_{\text{diff}}$, and serves as the fundamental driving force for the convergence of the contact point coordinates to the true position.

To determine the weight for compensating the residual into the predicted state, the optimal Kalman gain $\mathbf{K}[k]$ is solved:
\begin{equation}
    \mathbf{K}[k] = \mathbf{P}^{-}[k] \mathbf{H}[k]^T (\mathbf{H}[k] \mathbf{P}^{-}[k] \mathbf{H}[k]^T + \mathbf{M})^{-1}. \label{eq:kalman_gain}
\end{equation}
$\mathbf{K}[k]$ is an adaptive weight allocation operator: when the measurement noise $\mathbf{M}$ is much larger than the prediction covariance, the norm of $\mathbf{K}[k]$ decreases, and the system tends to maintain the predicted trajectory; when high-frequency, high-SNR angular velocity signals are generated immediately following the accelerometer trigger, the dynamic enhancement of $\mathbf{K}[k]$ significantly increases the correction component for $\mathbf{p}_c$, thereby accelerating convergence. This mechanism is analogous to the learning rate in gradient descent methods, but $\mathbf{K}[k]$ can adaptively adjust the update step size based on the uncertainty of the current state, demonstrating the advantage of the recursive Bayesian framework over fixed-step gradient descent.

Subsequently, the gain matrix is used to perform a posterior update of the predicted state, yielding the optimal estimate $\hat{\mathbf{x}}^{+}[k]$ at the current time:
\begin{equation}
    \hat{\mathbf{x}}^{+}[k] = \hat{\mathbf{x}}^{-}[k] + \mathbf{K}[k] \mathbf{y}[k]. \label{eq:posterior_update}
\end{equation}
This step achieves closed-loop error compensation in the state space. Since the angular velocity $\boldsymbol{\omega}[k]$ appears explicitly in the Jacobian matrix, a faster rotational speed of the manipulator leads to a more statistically significant correction step per unit residual. Finally, the state error covariance matrix is updated:
\begin{equation}
    \mathbf{P}^{+}[k] = (\mathbf{I} - \mathbf{K}[k] \mathbf{H}[k]) \mathbf{P}^{-}[k]. \label{eq:cov_update}
\end{equation}
This formula describes how the uncertainty in contact point position estimation is reduced by fusing new observation information. In practical experiments, the rapid decay of the trace of the $\mathbf{P}$ matrix is the core indicator of algorithm convergence. Due to the closed-form analytical Jacobian of the proposed model, a single iteration is computationally efficient on embedded platforms, supporting perceptual feedback above 1\,kHz, thereby providing real-time differential constraint information for subsequent trajectory planning.

For the planar probing tasks in this study, kinematic interactions and fingertip deformations are strictly confined to a 2D operational plane. Since out-of-plane gyroscope measurements consist primarily of negligible vibrations and noise, we project the 3D differential kinematic model onto this 2D plane. This dimensional reduction effectively eliminates unobservable degrees of freedom, prevents the accumulation of out-of-plane noise, and enhances numerical stability. Consistent with this 2D strategy, our experimental \gls{ekf} implementation operates at a discrete update frequency of 7~kHz. The process noise covariance is configured as a block-diagonal matrix $\mathbf{Q} = \text{diag}(\mathbf{Q}_p, \mathbf{Q}_v, \mathbf{Q}_c)$. 
Based on empirical tuning, the sub-matrices for the planar state variables are set to $\mathbf{Q}_p = \alpha_p\mathbf{I}_{2}$ for the \gls{tcp} position, $\mathbf{Q}_v = \alpha_q\mathbf{I}_{2}$ for the \gls{tcp} velocity, and $\mathbf{Q}_c = \alpha_c\mathbf{I}_{2}$ for the contact point coordinates, where $\mathbf{I}_{2}$ denotes the $2 \times 2$ identity matrix. The measurement noise covariance is defined as $\mathbf{M} = \text{diag}(\mathbf{M}_p, \mathbf{M}_v)$, with $\mathbf{M}_p = \beta_p\mathbf{I}_{2}$ representing the positional uncertainty of the robot's forward kinematics, and $\mathbf{M}_v = \beta_v\mathbf{I}_{2}$ reflecting the equivalent noise variance of the velocity observation. 
For all experiments in the following sections, we maintain constant parameter settings, specifically, $\alpha_p = 10^{-4}, \alpha_q = 10^{-3}, \alpha_c = 10^{-12}, \beta_p = 10^{-5},$ and $ \beta_v = 10^{-5}$.


\subsection{Angular Velocity Reconstruction and Calibration}

In the construction of the gyroscope-based tactile sensing framework, the raw angular velocity signals, denoted as $\boldsymbol{\omega}[k] = [\omega_x, \omega_y, \omega_z]^T$, captured by the embedded sensor inevitably exhibit physical attenuation and phase lag. This discrepancy stems from two coupled intrinsic mechanisms: the bulk viscoelastic deformation of the fingertip encapsulation medium (e.g., silicone) and the localized micro-slip at the contact interface. Ideally, a perfectly rigid and bonded tactile interface would ensure synchronous rotation between the sensor and the grasped object. In reality, the compliance and interfacial sliding cause the sensor’s local rotation ($\boldsymbol{\omega}$) to deviate from the object’s true rotation ($\tilde{\boldsymbol{\omega}}$). To tractably bridge this gap during interaction transients without solving intractable non-linear contact equations, we define an \textit{apparent transmission coefficient} $\eta_{app}$ to formulate a lumped observation operator:
\begin{equation}
    \tilde{\boldsymbol{\omega}} = \eta_{app} \cdot \boldsymbol{\omega} \label{eq:correction_factor}
\end{equation}
The physical validity of \cref{eq:correction_factor} is strictly governed by the contact state boundary conditions. Within the \textit{stick regime}, $\eta_{app}$ primarily reflects the bulk strain transmission efficiency dictated by the silicone’s viscoelastic compliance. As the tangential torque increases, the interface enters the \textit{micro-slip regime}, where localized boundary sliding manifests, causing $\eta_{app}$ to monotonically elevate. However, upon transitioning into the \textit{gross-slip regime} (macroscopic scaling), the rigid-body kinematics of the object completely decouple from the boundary displacement field of the fingertip, defining the ultimate physical boundary ($\eta_{app} \rightarrow \infty$) where this linear apparent mapping breaks down.

\begin{figure}[t]
    \centering
    \includegraphics[width=40mm]{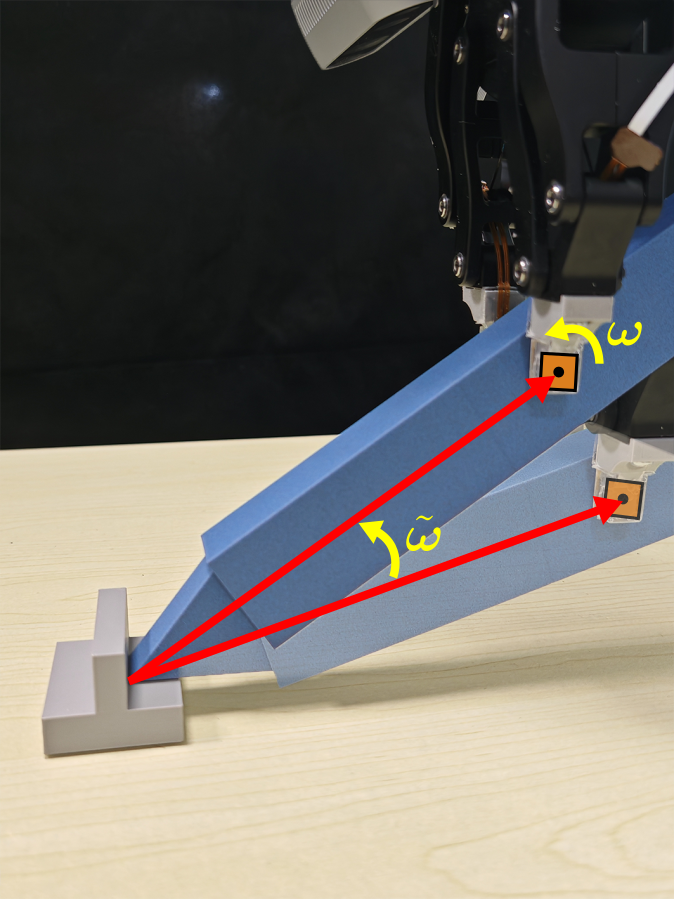}
    \caption{Schematic of the angular velocity discrepancy during contact-induced rotation. Theoretically, the embedded sensor and the object should share the same rotational velocity. However, due to the bulk viscoelastic deformation of the compliant fingertip and interfacial micro-slip, the raw angular velocity $\boldsymbol{\omega}$ measured by the sensor is attenuated compared to the true rotational angular velocity $\tilde{\boldsymbol{\omega}}$ of the grasped object, necessitating a physical compensation via the apparent transmission coefficient $\eta_{app}$.}
    \label{fig:kinematic_relation}
\end{figure}

The determination logic of the apparent coefficient $\eta_{app}$ is strictly founded on rigid-body kinematic constraints under the pre-macroscopic-slip boundary condition. According to classical kinematics, when a controlled object makes stable contact with an environment and induces a rotational tendency, a deterministic differential mapping exists among the instantaneous center of rotation $\mathbf{p}_c$, the spatial angular velocity $\tilde{\boldsymbol{\omega}}(t)$, and the linear velocity of the robotic end-effector $\mathbf{v}_g(t)$. This physical geometric consistency provides the natural optimization criterion for extracting $\eta_{app}$. To ensure the validity of this kinematic closure during calibration, the experiments are purposefully conducted within a load envelope that avoids gross slip, thereby establishing $\eta_{app}$ as a comprehensive medium transmission operator that encapsulates steady-state interfacial micro-slip and bulk elastomeric shearing, rather than a mere empirical correction term.

\begin{figure}[t]
    \centering
    \includegraphics[width=84mm]{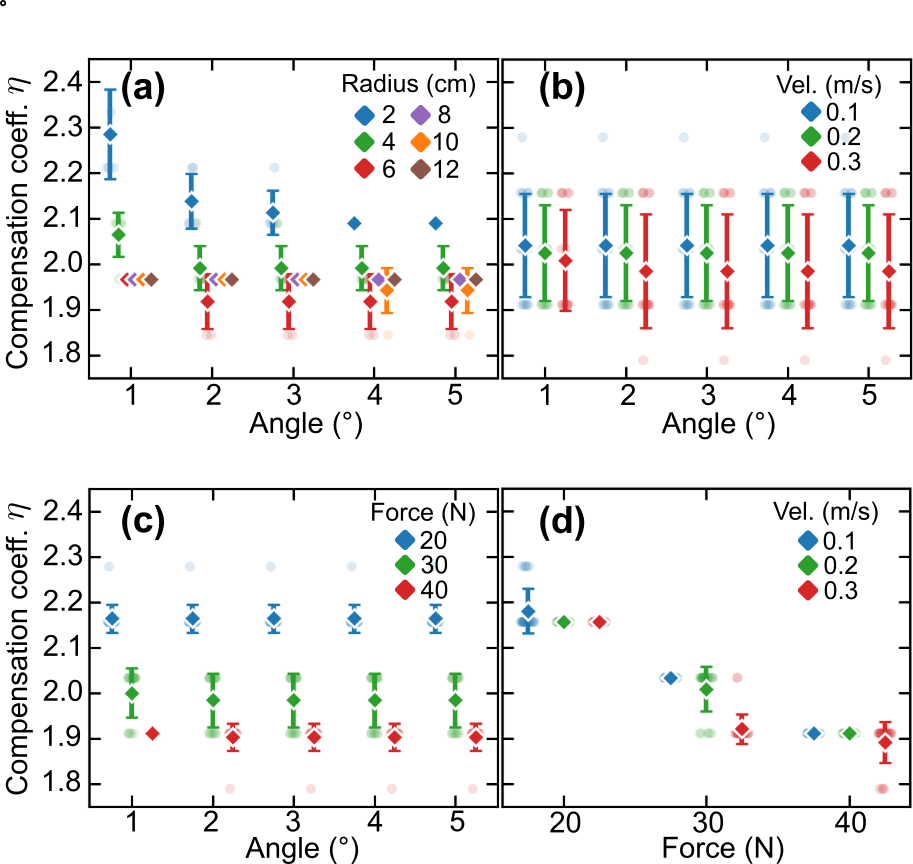}
    \caption{Statistical distribution and physical dependency analysis of the apparent transmission coefficient $\eta_{app}$. The subfigures illustrate the impact of (a) rotation radius, (b) probing speed, and (c) grasping force on $\eta_{app}$ under identical rotational deformation constraints, and (d) the cross-evaluation of force and speed. Individual raw data points are plotted as semi-transparent scatter points with horizontal jitter to prevent overlapping. Solid diamonds and vertical error bars denote the mean and \gls{sd} of each group, respectively.}
    \label{fig:calibration}
\end{figure}

To systematically calibrate $\eta_{app}$ and reveal its underlying mechanical dependencies, multi-dimensional controlled experiments were conducted. During contact-induced rotation, the torsional deformation limit of the silicone fingertip constitutes the essential physical constraint of the system observation. Therefore, the calibration experiments uniformly adopt the absolute rotation angle of the object as the independent variable to accurately evaluate the specific impact of different interaction conditions on the angular velocity transmission efficiency under identical deformation constraints.

As illustrated in \cref{fig:calibration}(a), the distance between the grasping position and the contact point strongly influences the apparent coefficient at close proximity, particularly below 4\,cm. Presumably, shorter rotation radii amplify the system’s sensitivity to minute displacements, thereby inducing more pronounced fluctuations in the $\eta_{app}$ distribution. Conversely, as the radius increases, the amplification effect of the geometric lever arm stabilizes the rotational transmission.
Furthermore, the results in \cref{fig:calibration}(b) confirm that the probing speed has a minimal impact on $\eta_{app}$ within the tested interaction frequency range. From the perspective of polymer mechanics, although silicone rubber exhibits inherent viscoelasticity, its loss tangent and storage modulus remain relatively stable under low-frequency, quasi-static cyclic loading. This strain-rate insensitivity within our operating regime ensures that the viscoelastic hysteresis does not introduce dominant non-linear artifacts, thereby providing strong physical support for adopting the linearized approximation in \cref{eq:correction_factor}.

In contrast, the grasping force notably modulates the apparent transmission coefficient, serving as the dominant governing factor. As shown in \cref{fig:calibration}(c), the value of $\eta_{app}$ exhibits a clear trend approaching $1.0$ as the grasping force scales up. Mechanically, this behavior cannot be simply attributed to the absolute elimination of interfacial slips; rather, it represents a transition toward a highly deterministic and uniform strain transmission state driven by dual bulk and boundary mechanisms. On one hand, an elevated normal force induces substantial volumetric compression within the silicone matrix, activating its strain-hardening regime and enhancing the local equivalent tangential stiffness. On the other hand, according to soft contact mechanics, a higher grasping force expands the central stick zone while constraining micro-slip to a stable, narrow peripheral annulus. Crucially, the intensified frictional coupling effectively suppresses the stochastic, chaotic stick-slip oscillations that typically distort boundary signals under light loads. Instead, the persistent micro-slip under high normal load operates in a highly continuous and regulated manner, allowing the tangential velocity field to be transmitted to the embedded rigid IMU with maximum fidelity and minimum phase distortion. This anchoring and stabilization effect synergistically drives the apparent velocity ratio $\eta_{app}$ toward unity, guaranteeing the statistical uniformity and linearity of the tactile perception framework.

The cross-evaluation of force and speed in \cref{fig:calibration}(d) further supports this conclusion, demonstrating that grasping force dominates the transmission efficiency and slip boundaries. Consequently, to handle varying operational loads, the coefficient $\eta_{app}$ is dynamically scheduled as a function of the real-time grasping force to sustain accurate state estimation. Through this physics-constrained modeling and analysis approach, the system effectively clarifies the interference mechanisms of complex external conditions. This ensures that in the subsequent \gls{ekf}-based state estimation, the system consistently operates within an observation framework compliant with rigid-body kinematic laws on the \gls{imu} by utilizing $\tilde{\boldsymbol{\omega}}(t)$, thereby achieving high-precision real-time tracking of contact points in complex occluded environments. Unless otherwise specified, all subsequent experiments adopt a grasping force of 30\,N and a corresponding baseline compensation coefficient of $\eta_{app} = 2.0$.


\section{Simulation}
Object pick-and-place tasks in unstructured environments are characterized by frequent collisions between the grasped object and its surroundings. In these scenarios, line contact emerges as a prevalent interaction modality, given that many everyday objects exhibit sharp, straight edges.
Such interactions often induce rotations of grasped object around the contact edge, effectively constraining the motion to a single rotational axis (\cref{fig:modeling}). We propose a framework that enables the robot to leverage \gls{imu} measurements of fingertip twisting and kinematic constraints to estimate contact edge positions.
To evaluate the performance of the proposed framework in line-contact scenarios, we developed a two-dimensional planar physics simulation with a geometric configuration that aligns with the experimental setup illustrated in \cref{fig:modeling}. The simulation replicates the dynamic process of a parallel gripper holding a rigid body making extrinsic contact with the environmental surroundings.

\subsection{Contact Position Estimation by Translational Probing}
\begin{algorithm}[h]
\caption{EKF-based Contact Point Estimation}
\label{alg:ekf_process}
\begin{algorithmic}[1]
\Require Kinematic measurements $\mathbf{z}[k] \equiv [\mathbf{p}_g^T[k], \mathbf{v}_g^T[k]]^T$, compensated gyroscope reading $\tilde{\boldsymbol{\omega}}[k]$, initial state $\hat{\mathbf{x}}[0]$, covariances $\mathbf{P}[0], \mathbf{Q}, \mathbf{M}$
\Ensure Estimated contact position $\hat{\mathbf{p}}_c$

\State Initialize state $\hat{\mathbf{x}}[0] \leftarrow [\mathbf{z}^T[0], \mathbf{0}^T]^T$ and $\mathbf{P}[0]$
\For{each time step $k$}
    \State $\hat{\mathbf{x}}^{-}[k], \mathbf{P}^{-}[k] \leftarrow \text{Predict}(\hat{\mathbf{x}}[k-1], \mathbf{P}[k-1], \tilde{\boldsymbol{\omega}}[k], \mathbf{Q})$ 
    
    \State $\mathbf{y}[k] \leftarrow \mathbf{z}[k] - h(\hat{\mathbf{x}}^{-}[k])$ 
    \State $\mathbf{H}[k] \leftarrow \partial h / \partial \mathbf{x} |_{\hat{\mathbf{x}}^{-}[k]}$ 
    \State $\hat{\mathbf{x}}[k], \mathbf{P}[k] \leftarrow \text{Update}(\hat{\mathbf{x}}^{-}[k], \mathbf{P}^{-}[k], \mathbf{y}[k], \mathbf{H}[k], \mathbf{M})$ 
\EndFor

\State \Return $\hat{\mathbf{p}}_c \leftarrow \text{ExtractPosition}(\hat{\mathbf{x}}[k])$
\end{algorithmic}
\end{algorithm}
To evaluate the performance of \cref{alg:ekf_process} under controlled conditions, we developed a 2D planar physics simulation. 
The environment replicates the geometric configuration shown in \cref{fig:modeling}, where a parallel gripper holds a rigid body making line contact with a fixed boundary. To simulate realistic sensing conditions, the \gls{imu} and \gls{tcp} data streams are corrupted with additive white Gaussian noise ($\sigma_\omega = 0.01\,\text{rad/s}$, $\sigma_p = 0.5\,\text{mm}$), and the sampling frequencies are set to $7\,\text{kHz}$ and $65\,\text{Hz}$ respectively, aligning with our physical hardware.

As detailed in \cref{alg:ekf_process}, the framework recursively estimates the contact position by fusing these high-frequency measurements. During this probing phase, the gripper is commanded to perform a linear translation orthogonal to the rotation radius $L$. This specific motion direction is chosen to maximize the geometric observability, ensuring that the induced rotation provides the strongest possible signal for the \gls{ekf}. However, because the gripper pushes along a straight line while the object rotates about the contact edge, the distance $L$ changes continuously during probing. This leads to a systematic estimation drift that scales with the rotation angle.


\subsection{Trajectory Planning Based on Estimated Contact Position}\label{section:trajectoryplanning}
\begin{figure}[ht]
    \centering
    \includegraphics[width=78mm]{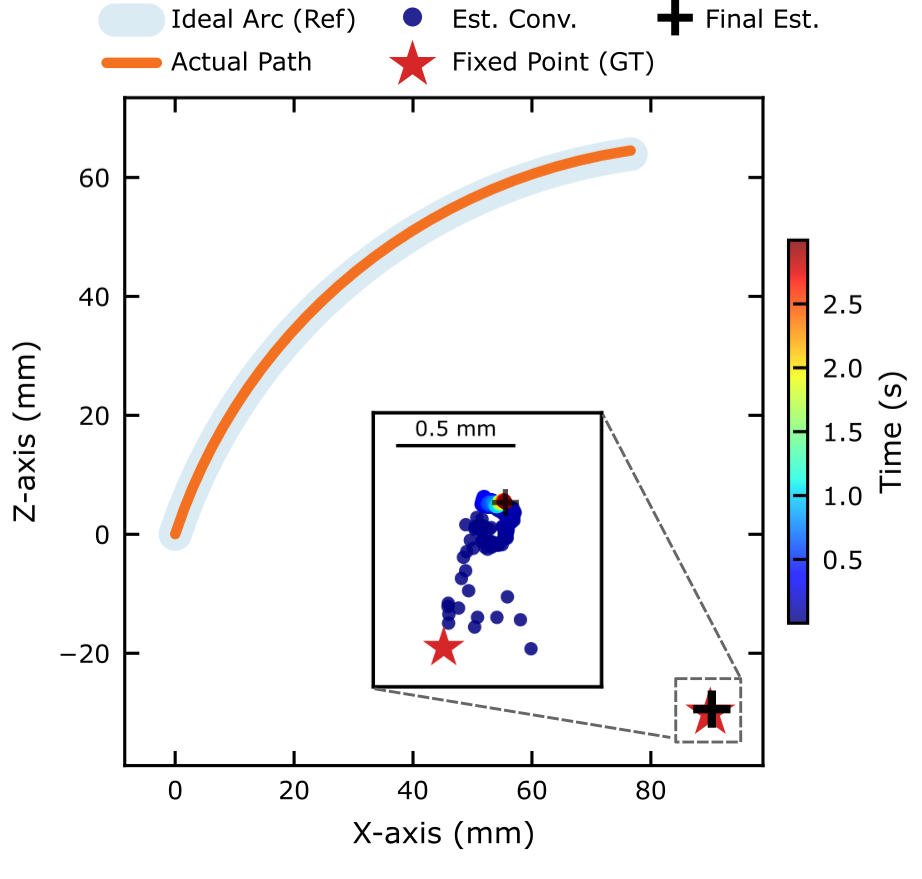}
    \caption{Closed-loop trajectory planning simulation based on the two-phase sensing strategy. The main plot illustrates the transition from initial translational probing (Phase 1) to kinematically consistent arc motion (Phase 2), where the actual executed path closely tracks the ideal reference. The magnified inset visualizes the dynamic convergence of the estimated contact point (color-coded by time gradient) toward the ground truth (red star).}
    \label{fig:simulation}
\end{figure}

To address the limitations of translational probing, our framework enables the robot to implement a kinematically consistent arc trajectory. Following a rapid initial estimation of the contact edge, the robot adapts its path to match the object's rotational arc, facilitating a longer window for data acquisition and more precise state estimation.
Beyond extending the estimation time window, this approach can also enable the robot to plan trajectories that rotate the grasped object about a fixed contact edge.
This is achieved without visual feedback or a priori geometric knowledge of the contact edge, relying exclusively on the closed-loop control leveraging the fusion of tactile (gripper tip \gls{imu}) and proprioceptive (robot \gls{tcp}) data.
The proposed twin-phase closed-loop control for trajectory planning is detailed in \cref{alg:two_phase_sensing} and is validated with simulation.

\begin{algorithm}[ht]
\caption{Two-Phase Active Sensing and Closed-Loop Planning Scheme}
\label{alg:two_phase_sensing}
\begin{algorithmic}[1]
\Require True contact point $\mathbf{p}_c^*$, initial gripper position $\mathbf{p}_g[0]$, sampling period $\Delta t$, total steps $N$, probing steps $K$ ($K<N$); process noise $\mathbf{Q}$, measurement noise $\mathbf{M}$, initial covariance $\mathbf{P}[0]$; noise std $\sigma_p$, $\sigma_\omega$
\Ensure Estimated contact position $\hat{\mathbf{p}}_c$

\Statex \textit{// Phase 1: Open-loop tangential probing (fixed direction)}
\State Compute initial radius vector $\mathbf{r}_0 = \mathbf{p}_g[0] - \mathbf{p}_c^*$
\State Compute tangential unit vector $\hat{\mathbf{t}} \perp \mathbf{r}_0$
\State Initialize EKF state $\hat{\mathbf{x}}[0] \leftarrow [\mathbf{p}_g^T[0], \mathbf{0}^T, \mathbf{0}^T]^T$, covariance $\mathbf{P}[0]$
\For{$k = 1$ to $K$}
    \State $\Delta s \leftarrow$ small step increment (e.g., linear in $k$)
    \State $\Delta \mathbf{p}_{\text{cmd}} \leftarrow \Delta s \cdot \hat{\mathbf{t}}$
    \State $\Delta \mathbf{p}_{\text{meas}}[k] \leftarrow \Delta \mathbf{p}_{\text{cmd}} + \mathcal{N}(0, \sigma_p^2)$
    \State Obtain noisy gyroscope measurement $\tilde{\boldsymbol{\omega}}[k]$ from the physics simulator
    \State $\mathbf{p}_g[k] = \mathbf{p}_g[k-1] + \Delta \mathbf{p}_{\text{meas}}[k]$, $\mathbf{v}_g[k] = \Delta \mathbf{p}_{\text{meas}}[k] / \Delta t$
    \State $\hat{\mathbf{x}}[k], \mathbf{P}[k] \leftarrow \text{EKF}\big(\hat{\mathbf{x}}[k-1], \mathbf{P}[k-1], \mathbf{p}_g[k], \mathbf{v}_g[k], \tilde{\boldsymbol{\omega}}[k], \mathbf{Q}, \mathbf{M}\big)$
\EndFor

\Statex \textit{// Phase 2: Closed-loop control (tangential direction updated online)}
\For{$k = K+1$ to $N$}
    \State Extract $\hat{\mathbf{p}}_c[k-1]$ from $\hat{\mathbf{x}}[k-1]$
    \State Obtain gyro measurement $\tilde{\boldsymbol{\omega}}[k]$ (with added noise)
    \State $\Delta\theta \leftarrow \tilde{\omega}_y[k] \cdot \Delta t$
    \State Compute tangential command $\Delta \mathbf{p}_{\text{cmd}}$ using $\hat{\mathbf{p}}_c[k-1]$ and $\Delta\theta$ (rotation about estimated contact)
    \State $\Delta \mathbf{p}_{\text{meas}}[k] \leftarrow \Delta \mathbf{p}_{\text{cmd}} + \mathcal{N}(0, \sigma_p^2)$
    \State $\mathbf{p}_g[k] = \mathbf{p}_g[k-1] + \Delta \mathbf{p}_{\text{meas}}[k]$, $\mathbf{v}_g[k] = \Delta \mathbf{p}_{\text{meas}}[k] / \Delta t$
    \State $\hat{\mathbf{x}}[k], \mathbf{P}[k]\leftarrow\text{EKF}\big(\hat{\mathbf{x}}[k-1], \mathbf{P}[k-1], \mathbf{p}_g[k], \mathbf{v}_g[k], \tilde{\boldsymbol{\omega}}[k], 
    \mathbf{Q}, \mathbf{M}\big)$
\EndFor

\State \Return $\hat{\mathbf{p}}_c = \text{ExtractContact}(\hat{\mathbf{x}}[N])$
\end{algorithmic}
\end{algorithm}

In Phase~1, the system performs a brief translational probing over a displacement of 5\,mm. As visualized in the magnified inset of \cref{fig:simulation}, the earliest estimates (dark blue scatter points) lie very close to the ground truth (red star). However, because the \gls{ekf} assumes pure rotation about the contact point, the estimate progressively drifts as translation continues. The system transitions to Phase~2 after a small displacement.

In Phase~2, the system switches to closed-loop arc following. At each control cycle, the \gls{tcp} trajectory is replanned around the current estimate $\hat{\mathbf{p}}_c[k-1]$ using the measured angular velocity $\tilde{\boldsymbol{\omega}}[k]$ to compute a tangential displacement command. Under arc motion, the kinematics match the \gls{ekf}'s rotation model: the estimate stops drifting and stabilizes. The macroscopic trajectory plot shows that the executed path closely tracks the ideal reference arc, with tracking deviation confined to millimeter level.


\section{Experiment A: Exemplary Scene}
This section evaluates the proposed contact localization framework under controlled conditions using a physical robotic platform. To simulate everyday objects systematically, we employ 3D-printed \gls{pla} targets featuring standardized geometric primitives: a rectangular block for line contact, and a sharp probe tip and a plate with a hemispherical cavity to enforce point contact. This controlled setup allows each physical parameter to be isolated and its individual influence on estimation accuracy quantified. 

The evaluation comprises three tests. First, we evaluate the scenario where the grasped object establishes line contact under constrained gripper movement.
In this case, the grasped object is already engaged with an environmental edge and the gripper executes a steady, constrained rotation about that contact boundary. We systematically vary the rotation radius, grasping force, and probing velocity to characterize their individual effects on accuracy. Within this analysis, we compare the proposed differential-constraint EKF against a geometric-constraint gradient-descent baseline to isolate the benefit of avoiding angular velocity integration.
Second, we evaluate the system's ability to localize line contact under ongoing gripper motion. The robotic gripper translates the grasped object toward a fixed boundary at a constant approach speed, generating an impact event while maintaining the ongoing motion. This scenario simulates real-time contact detection and localization during continuous robot operation without prior knowledge of environmental obstacles.
Third, we extend the framework from planar (2D) line contact to 3D point contact, treating the latter as a vector superposition of orthogonal planar constraints.
Together, these experiments establish the baseline performance envelope of the system under conditions where the ground-truth of contact geometry and location is measurable.

\subsection{Line Contact under Constrained Gripper Movement}\label{section:constrained}
\begin{figure*}[t]
    \centering
    \includegraphics[width=178mm]{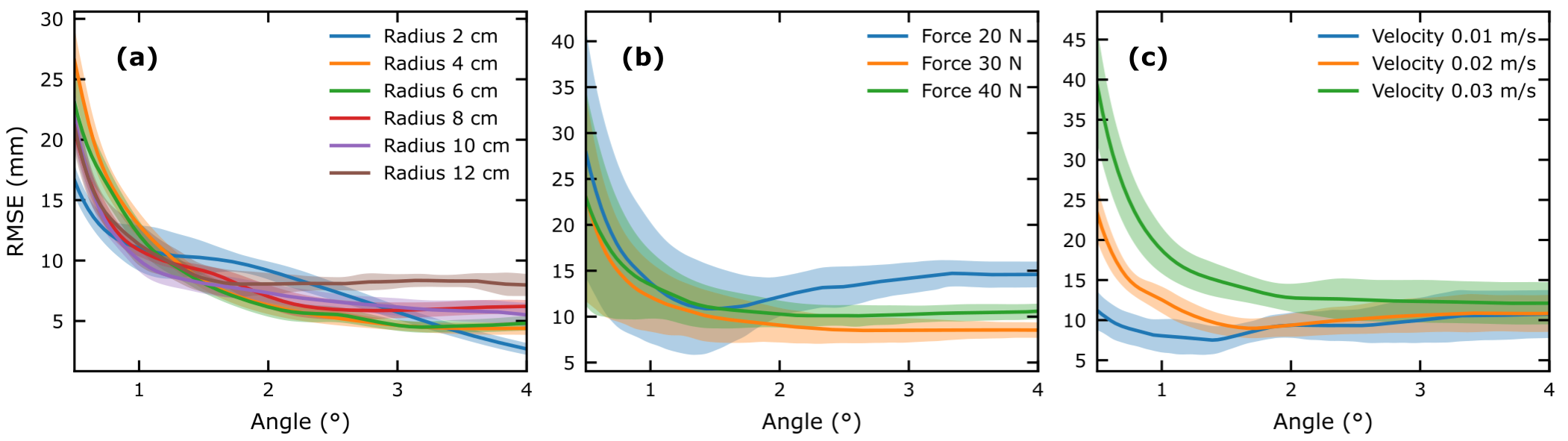} 
    \caption{Evaluation of localization accuracy (\gls{rmse}) under different experimental configurations. (a) Effect of rotation radius , under a fixed grasping force of 30~N and a probing velocity of 0.02m/s. (b) Effect of grasping force, under a fixed radius of 12cm and a probing velocity of 0.02m/s. (c) Effect of probing velocity, under a fixed radius of 12cm and a grasping force of 30N. Solid lines represent the mean error across 5 independent trials, and shaded areas indicate the standard deviation.}
    \label{fig:line_results}
\end{figure*}
To simulate the physical constraints between the object edge and the environment, a rectangular \gls{pla} block was 3D printed and used as the grasped object (\cref{fig:kinematic_relation}). 
This experiment employs only translational motion as the probing path. 
A constrained rotation of the object around the contact edge is naturally induced when moving the gripper in a direction loosely tangential to the rotating arc.

\subsubsection{Experimental Design and Execution}
To evaluate sensing performance and repeatability, we performed two experiments. Each experimental configuration was subjected to five trials to characterize the system's performance variability.
The first experiment characterizes the impact of the probing radius $L$, defined as the distance between the grasp contact and the rotation axis, on localization accuracy. We evaluated performance by scaling $L$ from 2 to 12\,cm in increments of 2\,cm.
In this group, the grasping force and probing velocity are fixed at 30~N (matching the compensation condition of $\eta = 2$) and 0.02~m/s, respectively, within an induced rotation range of $0.5^\circ$ to $4.0^\circ$. 
The second experiment investigates system stability via a full factorial design, examining the coupling effects across three levels of grasping force (20, 30, and 40\,N) and probing velocity (0.01, 0.02, and 0.03\,m/s). For these tests, the rotation radius is fixed at 12~cm. This design ensures that the influence of each parameter is evaluated under a consistent baseline of other variables.

\subsubsection{Accuracy Evaluation} 
As illustrated in \cref{fig:line_results}(a), the localization error decreases as the induced rotation angle increases. Notably, this trend reveals that in real-world scenarios, the estimation precision is primarily governed by the \gls{snr} at the onset of motion. A sufficient angular displacement provides the localization algorithm with enough accumulated gyroscopic measurements to overcome initial noise, driving the estimation to converge. 
Although the rotation radius $L$ exerts a secondary influence where longer rotation radii $L$ slightly limit final precision due to amplified end-effector jitter, the rotation angle remains the dominant factor for convergence.

The impact of grasping force, shown in \cref{fig:line_results}(b), indicates that a 30~N force yields the best performance for the current sensor configuration. This is because the compensation coefficient $\eta$ was calibrated specifically at this grasping force; when the force deviates from the calibration value, $\eta$ no longer accurately captures the current strain transmission characteristics of the encapsulating medium, leading to degraded estimation accuracy. Furthermore, the velocity analysis in \cref{fig:line_results}(c) demonstrates that low-speed probing (0.01~m/s) ensures a quasi-static interaction and superior stability. In contrast, high-speed probing (0.03~m/s) introduces significant transient deviations, as rapid impact induces high-frequency mechanical vibrations that exceed the modeling range of the current kinematic observer.

\subsubsection{Algorithm Comparison}
Based on the preceding parametric analysis, the rotation radius, grasping force, and probing velocity are fixed at $L = 12$\,cm, $F = 30$\,N, and $v = 0.02$\,m/s for the algorithm comparison and all subsequent experiments in this section.

Extrinsic contact localization was achieved by minimizing the least-squares loss in \cref{eq:geo_loss} using \gls{svd} and by tracking the in-finger rotation of a grasped object via visuotactile marker displacements \cite{ma2021extrinsic} . In contrast, this paper proposes solving the underlying differential constraints, as defined in \cref{eq:diff_loss}. Specifically, we directly track the twisting velocity of the fingertip via the \gls{imu} embedded inside and utilize Bayesian filtering to mitigate measurement noise.
\begin{figure}[ht]
    \centering
    \includegraphics[width=78mm]{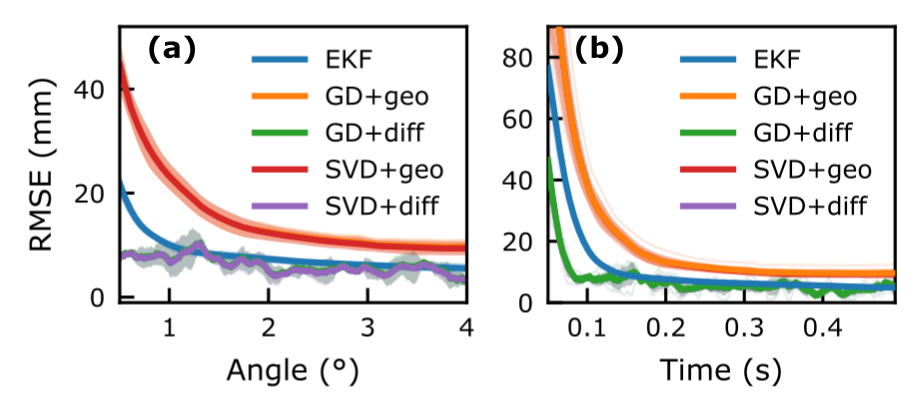}
    \caption{Comparison of five localization methods under real environmental noise. (a) \gls{rmse} as a function of induced rotation angle. (b) \gls{rmse} as a function of time since contact onset. Under the same constraint formulation, GD and SVD curves nearly coincide. Geometric-constraint methods yield the largest error, while the \gls{ekf} achieves smooth convergence comparable to the best accuracy of differential-constraint optimizers without their high-frequency oscillations.}
    \label{fig:ekf_vs_gd}
\end{figure}

To evaluate the proposed \gls{ekf} algorithm based on differential constraints, this section reproduces the \gls{svd} method used in \cite{ma2021extrinsic} and introduces a \gls{gd} method based on geometric constraints as a comparative baseline. All methods operate on identically pre-processed gyroscope signals, with static bias removed during initialization and high-frequency noise attenuated by a low-pass filter. In the following, we denote each method as solver+constraint, where geo and diff refer to the geometric (\cref{eq:geo_loss}) and differential (\cref{eq:diff_loss}) constraint formulations, respectively.

At each sampling time step, given the displacement increment and the integrated rotation matrix, the algorithm fits the geometric constraint by employing an optimizer to solve for the contact point coordinates that minimize the loss defined in \cref{eq:geo_loss}. To maintain trajectory continuity, the optimization framework initializes the current iteration using the estimated coordinates from the previous time step.

While the geometric constraint-based \gls{svd} and \gls{gd} methods can theoretically estimate the contact position, they exhibit significant limitations with real-world physical data. Constructing geometric constraints requires time integration of the gyroscope’s angular velocity signals to obtain the absolute rotation matrix. Despite the usage of the identical preprocessing method, residual low-frequency noise and bias accumulate during integration, introducing convergence lag: the optimizer must overcome progressively growing integration error before the estimate can stabilize. In contrast, the EKF-based framework proposed in this paper is fundamentally grounded in instantaneous differential constraints as shown in \cref{eq:diff_loss}. By relying solely on the instantaneous mapping between linear and angular velocities, it inherently circumvents the accumulation of integration errors.
Furthermore, we also applied the \gls{svd} and \gls{gd} optimizer directly to the differential constraint equations. However, experiments revealed that because instantaneous motion signals in real environments are affected by high-frequency noise and mechanical vibrations, pure numerical optimization lacking state priors is highly susceptible to local noise interference, resulting in severe oscillations or even divergence. 
The experimental results in \cref{fig:ekf_vs_gd} show that under the same constraint formulation, the GD and SVD curves nearly coincide, indicating that the constraint type, geometric versus differential, dominates over the solver choice. Geometric-constraint methods (GD+geo and SVD+geo) consistently yield the largest error, as expected from the integration drift discussed above. Among differential-constraint solvers, GD+diff and SVD+diff converge slightly faster than the \gls{ekf} initially and reach a comparable final accuracy, but both suffer from pronounced high-frequency oscillations. In contrast, the \gls{ekf} produces smooth estimates throughout, confirming that its recursive state prior acts as an effective regularizer against the instantaneous noise that destabilizes pure per-step optimization.


\subsection{Line Contact under Ongoing Gripper Movement}\label{section:ongoing} %
\begin{figure}[t]
    \centering
    \includegraphics[width=60mm]{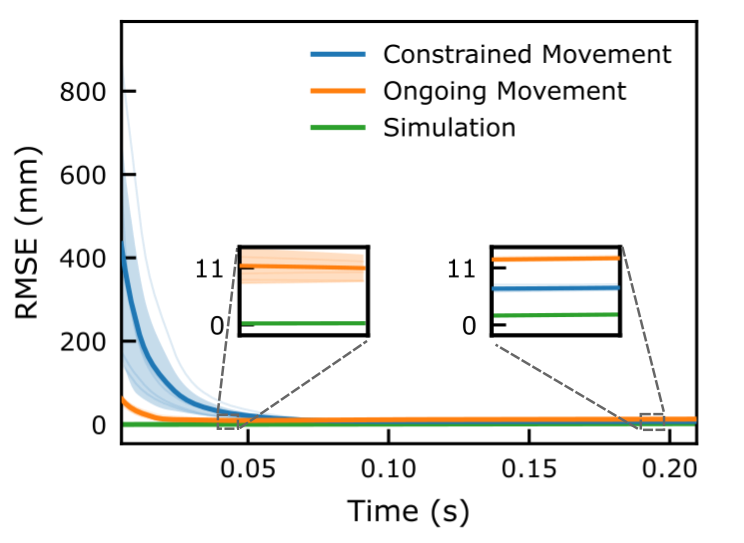}
    \caption{\gls{rmse} comparison of the proposed \gls{ekf} under constrained-gripper and ongoing-movement probing conditions, with noise-free simulation included as a baseline reference. The zoomed inset details the converged region, revealing the trade-off between convergence speed and steady-state accuracy across the two probing regimes.}
    \label{fig:static_vs_dynamic}
\end{figure}
While the previous scenario evaluated an object starting from an established line contact, many real-world tasks require a robot to manipulate an object through free space prior to making unexpected impact with the unstructured environment. Using the fixed parameter configuration established above, this experiment simulates the complete progression of an object approaching a contact boundary at a constant velocity. To precisely capture the transient mechanics of the physical interaction, the entire motion cycle is categorized into three distinct phases.

In the constant-speed approach phase, the robotic arm moves the object steadily at a preset velocity, during which the sensor recordings are limited to low-level system vibration noise. Upon physical interaction, the transient impact induces a constrained rotation of the object around the contact edge. The algorithm automatically identifies this contact moment by monitoring the transient spike in the accelerometer signals and extracts the corresponding data segment for EKF-based contact localization. Finally, in the braking phase, the system triggers a protective stop once the detected rotation angle reaches a preset threshold. 
This integrated design allows the data analysis to focus exclusively on the contact-rich period, directly demonstrating the tracking efficacy of the \gls{tecdar} framework within unmodeled and unstructured environments.

As shown in \cref{fig:static_vs_dynamic}, the ongoing probing process achieves a markedly faster convergence rate than the constrained-gripper configuration. At the moment of collision, the initial momentum of the grasped object generates an impact transient with high \gls{snr}, providing strong excitation for the differential constraints and driving rapid error reduction at contact onset. In addition, gyroscope measurements acquired during continuous motion exhibit a smoother envelope, since the system avoids the nonlinear friction transitions inherent in start-stop operation. 
The localization algorithm converges to an \gls{rmse} of 11\,mm within 30\,ms of impact, whereas the constrained-gripper configuration, starting from a stationary pose, yields an \gls{rmse} of 50\,mm at the same 30\,ms window.

For the constrained-movement scenario, the localization algorithm achieves a steady-state \gls{rmse} of 8\,mm, outperforming the 11\,mm error observed in the ongoing-movement case. This outcome highlights a distinct trade-off in estimation dynamics: while the high-SNR collision transient provides sufficient instantaneous information to rapidly drive the \gls{ekf} state toward the true contact location, the extended observation window during quasi-static probing allows the filter to integrate a larger volume of consistent measurements. Consequently, this progressive refinement yields a level of long-term accuracy beyond what the brief impact transient alone can support. 
These results demonstrate that the \gls{tecdar} framework enables real-time, detect-on-collision sensing during continuous robot motion, localizing environmental contacts at a millisecond scale without prior knowledge of obstacle positions.


\subsection{Point Contact}\label{section:pointcontact} 
\begin{figure}[ht]
    \centering
    \includegraphics[width=60mm]{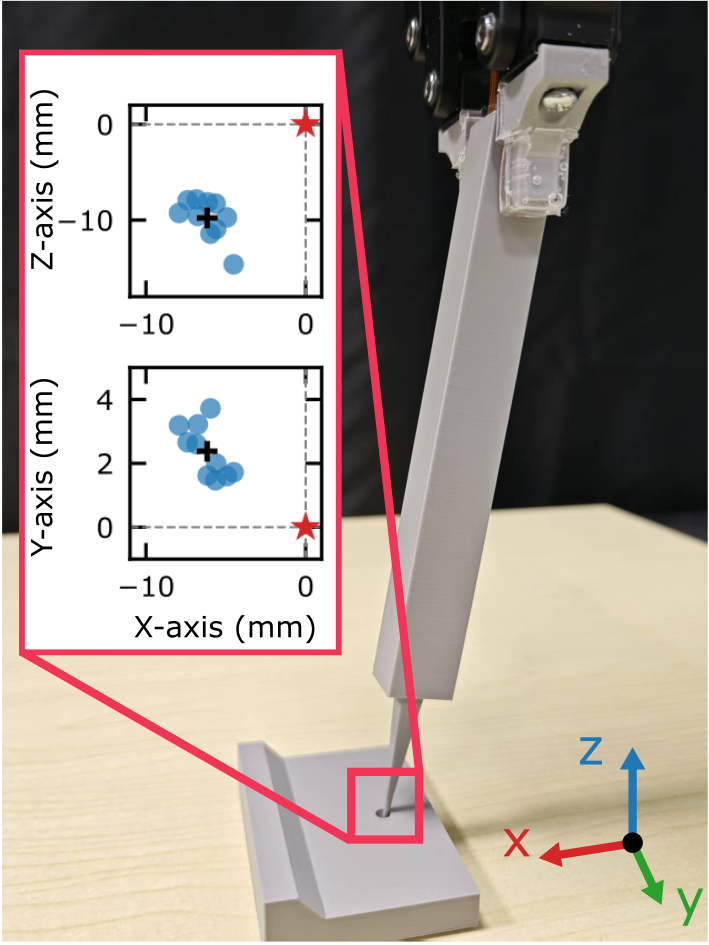}
    \caption{Experimental results of the point contact localization task. The scatter plot shows the distribution of the final estimated contact points across ten trials, illustrating the high repeatability of the localization algorithm in 3D space.}
    \label{fig:point_task}
\end{figure}
In addition to the scenarios of line contact, this study further explores the point contact case to verify the generalization capability of the sensing framework toward more complex geometric constraints. From a geometric kinematics perspective, stable point contact in 3D space can be viewed as the vector superposition of line contact constraints within two mutually orthogonal planes. The experiment utilizes a probing tool of length 140\,mm with a marker fixed at the end, interacting with a base plate with a $3$\,mm diameter cavity. During the probing process, the robotic arm drives the object to execute a circular arc trial motion in the horizontal plane, which fully activates multi-dimensional observation information by continuously changing the direction of the contact vector to resolve the 3D coordinates.

As illustrated in \cref{fig:point_task}, the final estimated points from ten repeated experiments are tightly concentrated. The per-axis estimation errors, expressed as mean $\pm$ standard deviation, are $-4.17 \pm 1.01$\,mm ($x$-axis), $2.38 \pm 0.77$\,mm ($y$-axis), and $-9.77 \pm 2.00$\,mm ($z$-axis). The $z$-axis offset is the largest among the three, consistent with the in-plane circular probing motion providing stronger geometric excitation for the horizontal coordinates than for the depth dimension. These results validate the vector-superposition formulation of orthogonal planar constraints for point-contact localization and establish a controlled baseline for generalizing this capability to everyday grasped objects.


\section{Experiment B: Household Task Validation}

To evaluate the system's performance in practical, unstructured environments, this section builds upon the previous validations conducted on idealized 3D-printed \gls{pla} models and in simulation, scaling up the system's perception capabilities to real-world, everyday object manipulation tasks. Accordingly, three progressive real-world experiments are designed to evaluate three core capabilities: contact position estimation, trajectory planning, and environmental mapping with pure tactile exploration.

\subsection{Case 1: Pen (Point Contact)}

\begin{figure}[t]
    \centering
    \includegraphics[width=82mm]{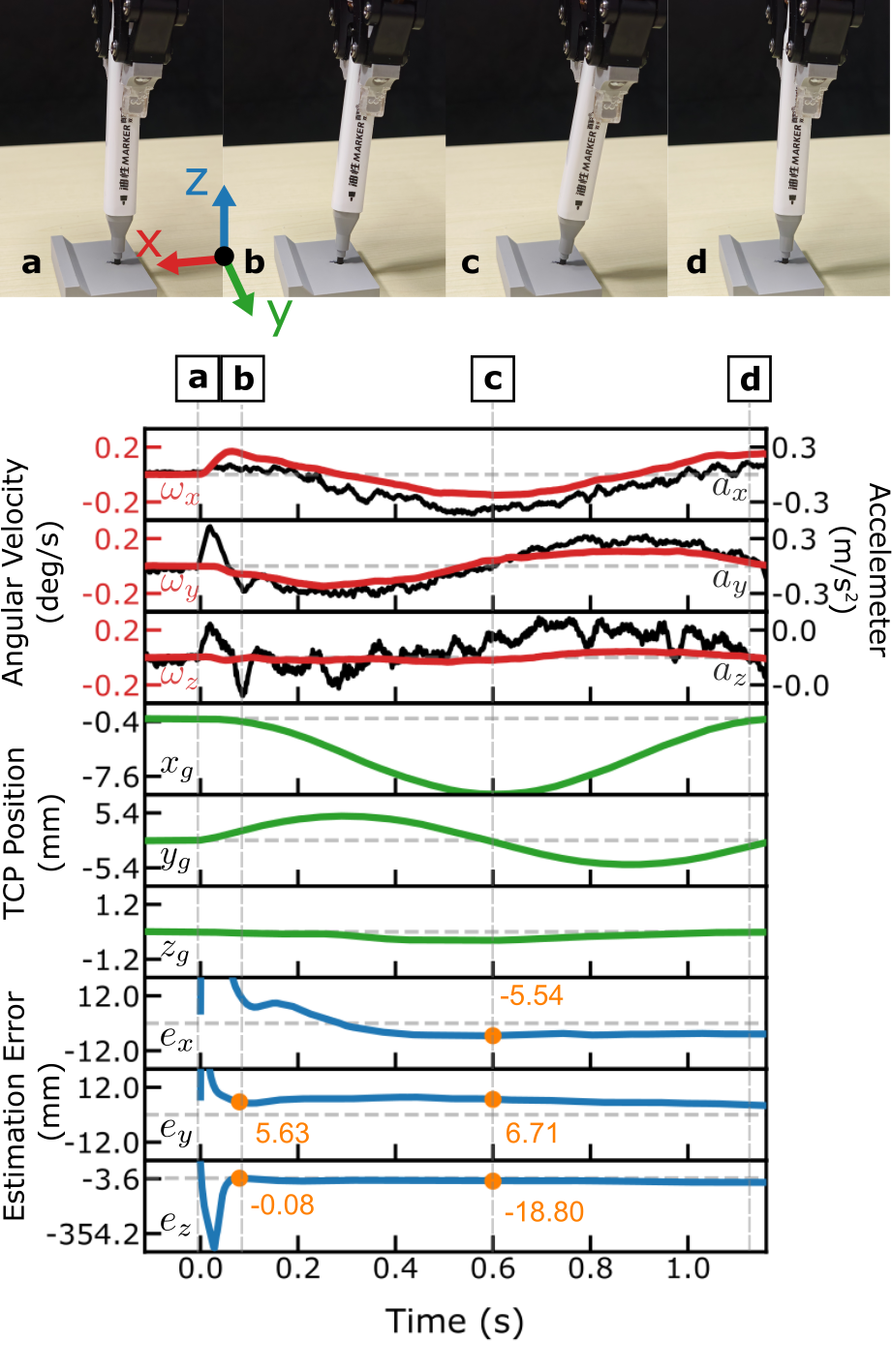} 
    \caption{Point contact experiment with a pen of length 90\,mm. The upper strobe sequence (a--d) shows the robotic arm holding the pen in contact with an external point while executing a full $360^\circ$ circular trajectory in the horizontal plane. The lower data plots illustrate the signal evolution: the red and black lines represent the gyroscope and accelerometer readings, respectively; the green line represents the \gls{tcp} displacement; and the blue line represents the estimation error relative to the ground-truth contact point. The error trace is absent before (a) because the localization algorithm is activated by the accelerometer contact trigger and does not run during the stationary phase.}
    \label{fig:real-world_pen}
\end{figure}

The first experiment tests how well the point-contact localization generalizes to everyday objects, using a marker pen of length 90\,mm as the grasped probing tool.
The experimental scenario simulates a common tool-handling process: the robotic arm grasps a pen as a general-purpose probing tool, bringing its sharp tip into interaction with an unknown fixed point in the external environment. The core of this task is to verify the algorithm's adaptability to real-world tools.
From a kinematic perspective, stable point contact in 3D space can be treated as the vector superposition of line-contact constraints within two mutually orthogonal planes, as discussed in \cref{section:pointcontact}. By executing a full $360^\circ$ circular trajectory that continuously varies the contact vector direction, the system fully activates multi-dimensional observation information to resolve the 3D coordinates of the contact point.

In this experiment, the robotic arm first commands the pen tip to maintain contact with the target point. Subsequently, the end-effector executes a preset small-amplitude circular trial motion of $360^\circ$ within the horizontal plane (\cref{fig:real-world_pen}). 
The gyroscope and accelerometer signals reveal distinct stages across the circular trajectory. At the initial stage (a), the pen is held stationary against the contact point, registering only baseline noise. Then, the end-effector initiates the circular motion, and the signals rise immediately. After around 80\,ms into the motion (b), the localization algorithm has already converged to a stable estimate, with per-axis errors of $5.63$\,mm ($y$-axis) and $-0.08$\,mm ($z$-axis), all averaged over three independent trials. 
The faster convergence along the $y$-axis between (a) and (b), relative to the $x$-axis which exhibits a quarter-phase rotational delay, is driven by the circular motion predominantly exciting the $\omega_x$ gyroscope component, whose richer observational information preferentially resolves the $y$- and $z$-axis coordinates.
(c) marks the halfway point of the trajectory ($180^\circ$, $0.6$\,s), at which the per-axis estimation errors are $-4.06$\,mm ($x$-axis), $6.11$\,mm ($y$-axis), and $-18.50$\,mm ($z$-axis).
The $z$-axis exhibits the largest error, consistent with prior finding that depth resolution being most sensitive to the instantaneous contact geometry during the circular trajectory. 


\subsection{Case 2: Paper Cutter (Trajectory Planning)} 
\begin{figure}[t]
    \centering
    \includegraphics[width=73mm]{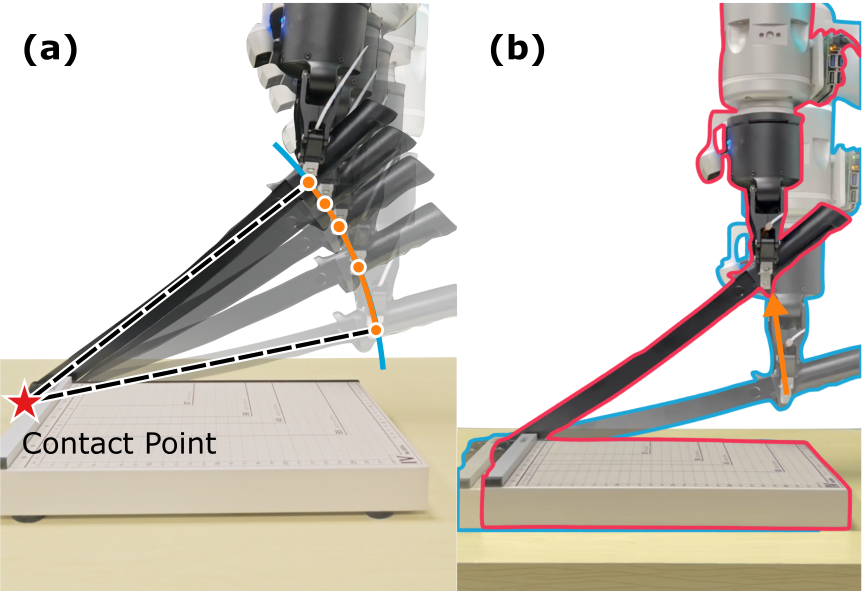}
    \caption{Stroboscopic trajectory of the paper cutter manipulation experiment. (a) Successful trial: the robot lifts the handle along a circular arc planned in real time based on the localization of the rotation axis; the cutter base remains stationary, and the strobe points closely track the true circular arc. (b) Failure case: the gripper moves along a fixed diagonal straight-line path of 15\,cm; the handle lifts but the cutter base displaces, violating the rotational constraint and causing the trajectory to deviate from the expected arc. }
    \label{fig:real-world_paper_cutter}
\end{figure}
This experiment evaluates the \gls{tecdar} framework for closed-loop trajectory planning in a real-world setup, utilizing \cref{alg:two_phase_sensing} detailed in \cref{section:trajectoryplanning}. Without any prior knowledge of the tool's geometry or hinge location, the robot must estimate the position of the rotational axis within a short time window immediately upon initiating tool motion. 

The real-world task involves opening and closing the blade of a paper cutter. The robot is programmed to move its gripper at a distance greater than 15\,cm.
As shown in \cref{fig:real-world_paper_cutter}a, the robot successfully completes the task by leveraging the sensing and planning framework enabled by \cref{alg:two_phase_sensing}, which maps the handle's trajectory based on the estimated rotational axis. Specifically, the system recursively fuses gyroscopic signals with manipulator displacement to localize the hinge, generating a real-time, constant-radius circular path. Because the planned trajectory accurately matches the cutter's kinematic radius, the cutter base remains completely stationary throughout the operation.

Conversely, when operating without the localization algorithm, the robot lacks prior knowledge of the hinge position. It consequently pulls the handle along a linear path, causing a severe forced displacement of the tool's base (\cref{fig:real-world_paper_cutter}b). This stark contrast demonstrates that estimating the rotational axis for trajectory planning is essential when manipulating articulated mechanisms that are not rigidly anchored to the environment.

\subsection{Case 3: Book (Line Contact)}\label{section:book}
\begin{figure*}[t]
    \centering
    \includegraphics[width=130mm]{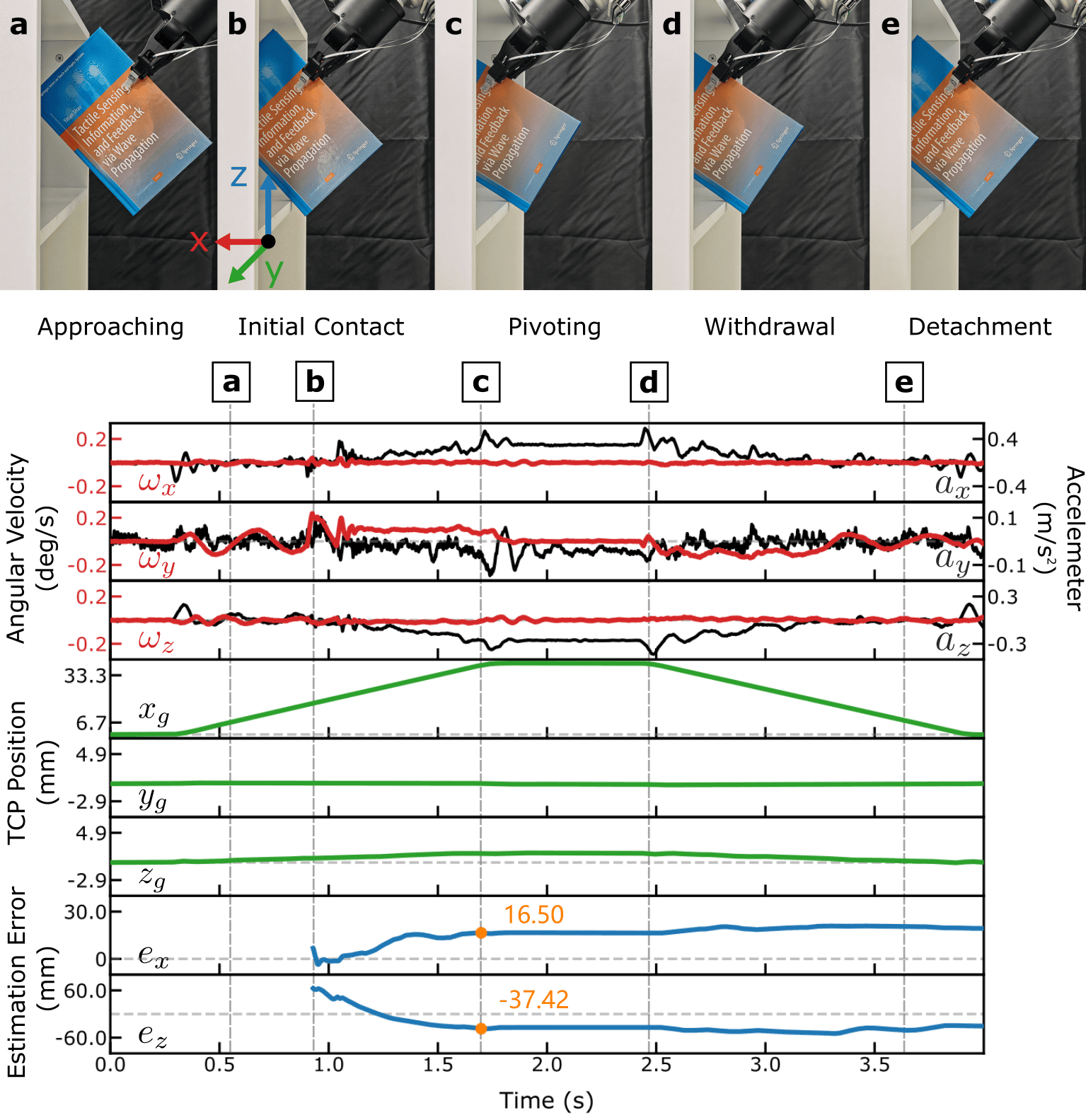} 
    \caption{Hardcover book line contact experiment. The upper strobe sequence (a--e) shows the complete task procedure: free-space transport toward the cabinet (a), collision with the cabinet frame (b), pivoting rotation about the contact edge (c), withdrawal along the insertion path (d), and full departure from the contact boundary (e). The lower data plots show the signal evolution: the red and black lines represent the gyroscope and accelerometer readings, respectively; the green line represents the \gls{tcp} displacement; and the blue line represents the estimation error relative to ground truth.}
    \label{fig:real-world_book}
\end{figure*}
Unexpected environmental collisions during the transportation of a grasped object induce a transient, edge-constrained rotation, providing geometric constraints allowing the robot to localize the contact edge.
Extending the exemplary ongoing line-contact evaluation (\cref{section:ongoing}), this experiment assesses \gls{tecdar}'s localization performance during ongoing gripper motion in real-world tasks.
The task scenario simulates a common household manipulation task: a robot holds a book and inserts it into a cabinet, during which the long, straight edge of the book establishes dynamic contact with the cabinet frame, imposing a typical line contact.

The complete task procedure and corresponding results are presented in \cref{fig:real-world_book}.
The robotic gripper transports the book through free space toward the cabinet at a constant velocity (a).
(b) captures the moment of sudden, unexpected collision between the book edge and the cabinet frame. Although the time-domain accelerometer transient is less prominent for the hardcover book than for the lighter probes used in the controlled experiments, contact onset is reliably detected by the spectral energy of the accelerometer signal in the 40--200\,Hz band. The localization algorithm is activated immediately upon detection.
Then, the robot continues driving the book into the cabinet, while the book edge undergoes a small-range pivoting rotation about the contact boundary on the cabinet frame (c).
The robot then withdraws the book by retracing its insertion path (d), returning to a fully departed state, thereby completing the perception cycle, as shown in (e).

Approximately 0.7\,s after initial contact, the robot localized the contact with estimation errors reaching 16.50 mm along the X-axis and -37.42 mm along the Z-axis, where they remain steady throughout the remainder of the insertion. These errors exceed those obtained in the controlled \gls{pla} block experiments (\cref{section:constrained}). The discrepancy is likely driven by the book's larger mass. During insertion into the cabinet, gravity pre-loads the soft PDMS fingertip, inducing a sustained shear deformation. When contact-induced torsion occurs, the resulting gyroscopic signals are superimposed onto this pre-existing shear state, compromising the rigid-body assumption that held true for the lightweight \gls{pla} block baseline.
Nevertheless, given the substantial 20\,cm rotation radius from the fingertip grasp to the contact edge, this absolute localization error translates to a relatively minor angular deviation. This indicates that the framework retains high practical utility even under the degraded signal conditions of real-world scenarios. 
Despite these non-ideal sensing effects, the \gls{tecdar} framework maintains a stable estimate across the entire insertion process, demonstrating the system's robustness in everyday manipulation tasks. Ultimately, this capability provides a reliable foundation for mapping large regions of unstructured environments, as demonstrated in the subsequent section.


\section{Experiment C: Real-Time Tactile Mapping of Unstructured Environment} 
\begin{figure*}[ht]
    \centering
    \includegraphics[width=120mm]{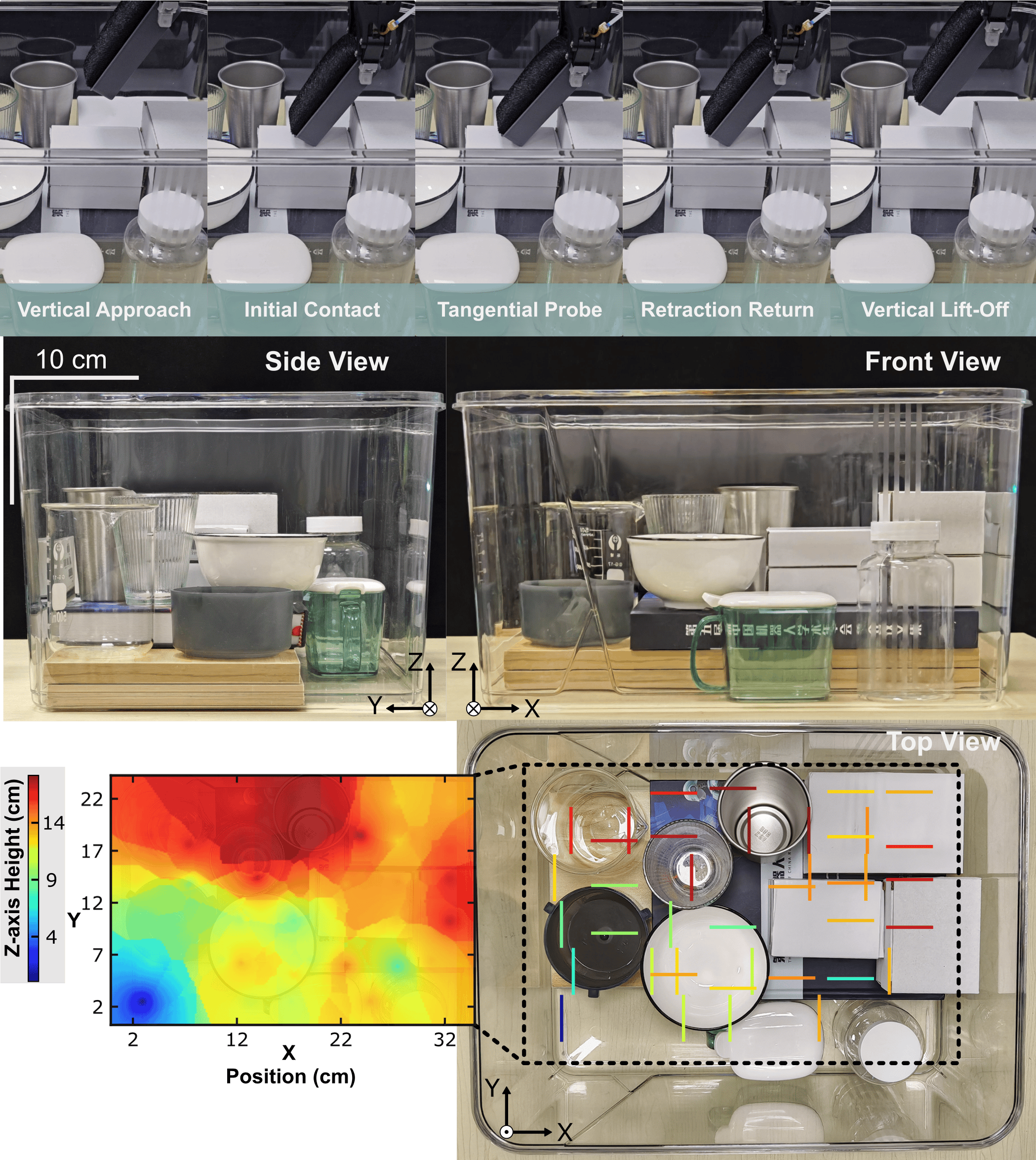}
    \caption{Top view of the grid scanning path with overlaid 3D reconstruction results in a real-world scene. 
\textbf{(Top)} The five-stage pipeline of the autonomous physical interaction probing consisting of approach, contact detection, tangential probing, retraction, and lift-off;
\textbf{(Middle)} side and front views of the complex experimental setup containing transparent and highly specular objects; 
\textbf{(Bottom-Right)} top view of the grid scanning together with the spatial projections of the localization results;
\textbf{(Bottom-Left)} spatial height heatmap reconstructed via Bayesian data fusion of the localization results; the colorbar indicating the relative height with the baseline platform offset removed.}
    \label{fig:demo_view}
\end{figure*} 
The evaluation in \cref{section:book} demonstrates the efficacy of the \gls{tecdar} framework in localizing unexpected line contacts during the free-space transportation of a grasped object. This capability can be further generalized to tactile exploration and mapping of an entire unstructured region relying solely on touch feedback.
To simulate a representative real-world scenario, our experimental setup houses a collection of everyday objects—including cups, bowls, books, and boxes—inside a transparent container with a size of 45 $\times$ 35 $\times$ 24.5\,cm (\cref{fig:demo_view}).
The reliability of conventional optical perception is fundamentally limited in such a scenario. Dynamic lighting, transparent or highly specular surfaces of glass and metal objects, and visual occlusions from the robotic manipulator or the surrounding environment heavily compromise vision-based 3D geometric mapping.
Hence, the proposed mapping framework operates entirely without visual input, relying solely on physical interactions to reconstruct the geometric layout of the unstructured environment within the container.
To demonstrate the versatility of the framework without loss of generality, the robot utilizes a common chalkboard eraser as an improvised probe to execute extrinsic contact localization.
The size of the eraser is 11 $\times$ 4.5 $\times$ 3.5\,cm.
The robot's parallel gripper holds the sides of the eraser with a constant force of 30\,N and performs vertical probing to explore the environment inside the container.

To map the entire container space, the system adopts a grid-based scanning strategy. Initiating the process at one corner of the container, the robot sequentially iterates through a discretized, evenly distributed $5\times6$ grid of sampling locations. 
Given that the robot utilizes the grasped eraser for vertical probing, the tool is secured in the gripper at a tilted angle of approximately $30^\circ$. This tilted orientation prevents direct, flat impacts against flat surface tops, thereby encouraging rotational movement of the eraser upon downward contact.
At each grid point, the robotic gripper executes a downward probing motion until the gripper \gls{imu} detects a collision between the grasped eraser and the environment. Upon reaching this contact threshold, the accelerometer (Z-axis ${>}\,0.4\,\text{m/s}^2$) and gyroscope (Y-axis ${>}\,0.3\,\text{rad/s}$) signals trigger an immediate state transition to the extrinsic contact localization phase, wherein the robot executes a trajectory loosely tangential to the eraser's rotational arc for a maximum distance of $5$\,mm.
During this movement, the \gls{tecdar} framework estimates the contact edge location in real time.
If the \gls{imu} fails to trigger the localization phase, such as when the eraser experiences a blunt collision against a flat surface, a secondary force-based threshold is applied. Once the robot's \gls{tcp} force exceeds $5\text{ N}$, the system aborts probing at the current grid location and immediately retracts as a safeguard measure.

Note that probing under a single pose can only obtain 2D coordinates within the rotational plane.
For instance, rotating the grasped object around the y-axis allows the \gls{tecdar} framework to estimate only the X-Z coordinates of the contact edge.
To achieve complete localization in 3D space, the system drives the robotic arm to rotate $90^\circ$ after completing the grid traversal, thereby aligning the rotation axis of the grasped object to x-axis and thereby estimating the Y-Z coordinates of the contact edge. 
Based on this new pose, the system executes a second full traversal over the same grid points. 
The 3D geometric profile of the explored environment can be estimated by fusing both traversal data, forming a Z-axis depth map over the X-Y plane.

The entire experimental procedure is shown in the top panel of \cref{fig:demo_view}. 
Each probing consists of five stages. 
During the approach phase (a), the robotic arm drives the gripper vertically downward. Upon detecting the surface impact via the inertial transient trigger (b), the system instantly transitions to the tangential probing phase (c), where the gripper moves orthogonally to the object's pose while estimating the contact location in real time. Finally, the tool retracts (d) and lifts off vertically (e) to reset for the next grid point. Exceeding the \gls{tcp} force threshold automatically bypasses (c), and the probing trial at that specific grid coordinate is marked as invalid.

As illustrated in the bottom-right of \cref{fig:demo_view}, the system projects the estimated contact edges from all valid probing trials into a unified spatial map overlaid on a top-view photo of the environment. We estimate the geometric profile of the container interior through Bayesian data fusion.

Let $\mathbf{x}_i \in \mathbb{R}^2$ and $d_i$ denote the 2D coordinate and estimated depth of the $i$-th probing point, for $i = 1, \dots, N$. Let $Z$ be the discrete random variable representing the depth at an arbitrary query location $\mathbf{x} \in \mathbb{R}^2$, with sample space $\mathcal{Z} = \{d_1, \dots, d_N\}$.
Without knowledge of $\mathbf{x}$, the prior $P[Z = d_i]$ is the probability of encountering depth $d_i$ at an arbitrary location in the environment.
The system initializes the first probing cycle with a uniform prior $P[Z = d_i] = N^{-1}$, which is replaced by an empirical prior if historical observations are available.
We model the likelihood of locating at $\mathbf{x}$ given measurement $d_i$ as a Gaussian kernel centered at the probing location $\mathbf{x}_i$:
\begin{equation}
P[\mathbf{x} \mid Z = d_i] \propto \exp\!\left(-\frac{\|\mathbf{x} - \mathbf{x}_i\|^2}{2\sigma^2}\right),
\label{eq:gaussian_likelihood}
\end{equation}
where $\sigma$ is the kernel bandwidth.
Applying Bayes' theorem:
\begin{align}
P[Z = d_i \mid \mathbf{x}]
&= \frac{P[\mathbf{x} \mid Z = d_i] \, P[Z = d_i]}{\sum_{j=1}^{N} P[\mathbf{x} \mid Z = d_j] \, P[Z = d_j]}
\label{eq:posterior_uniform}
\end{align} 
The second equality follows from substituting~\cref{eq:gaussian_likelihood} under a uniform prior.

Thus, the posterior expectation of the depth at $\mathbf{x}$ is
\begin{equation}
E[Z \mid \mathbf{x}] = \sum_{i=1}^{N} d_i \, P[Z = d_i \mid \mathbf{x}].
\label{eq:posterior_expectation}
\end{equation}
Substituting~\cref{eq:gaussian_likelihood} and \cref{eq:posterior_uniform} into~\cref{eq:posterior_expectation} and restricting the sum to the $k$ nearest neighbors of $\mathbf{x}$, since the Gaussian kernel decays exponentially with distance, yields the kernel-weighted depth estimator:
\begin{equation}
\hat{d}(\mathbf{x}) = \frac{
    \sum_{i \in \mathcal{N}_k(\mathbf{x})} d_i \,
    \exp\!\left(-\frac{\|\mathbf{x} - \mathbf{x}_i\|^2}{2\sigma^2}\right)
}{
    \sum_{i \in \mathcal{N}_k(\mathbf{x})}
    \exp\!\left(-\frac{\|\mathbf{x} - \mathbf{x}_i\|^2}{2\sigma^2}\right)
},
\label{eq:kernel_estimator}
\end{equation}
where $\mathcal{N}_k(\mathbf{x})$ denotes the set of $k$ nearest probing points to $\mathbf{x}$.

In our implementation, we set $k = 4$ and $\sigma = 2.2$. The resulting heatmap in~\cref{fig:demo_view} visualizes the posterior depth estimate $E[Z \mid \mathbf{x}]$ across the container, transforming discrete probe measurements into a continuous depth field. This heatmap reveals the 3D profiles and boundaries of the items inside the container, separating protruding beaker walls from the recessed container floor, and demonstrates that multi-point tactile probing can reconstruct macrogeometric features under severe visual degradation.


\section{Limitations}

Several aspects of the current framework warrant further discussion.
First, the kinematic model assumes rigid-body motion of the grasped object. Compliant or deformable objects would violate this premise; accordingly, the present experiments are limited to rigid blocks and everyday objects with sufficient stiffness (such as pen, paper cutter, book, chalkboard eraser). Performance on soft or irregularly shaped objects remains unexplored.

Second, \cref{alg:ekf_process} relies on the kinematic constraint imposed by a stable contact between object and environment. Slip at this interface, whether from low surface friction, surface irregularity, or impact-induced rebound, breaks this constraint; the corresponding effect on localization accuracy is yet to be characterized quantitatively. 
Moreover, under constrained movement, a related failure mode occurs when the probing velocity vector points toward the rotation axis, which fails to generate sufficient rotational excitation for gyroscopic sensing, which results in low-SNR signals and slower convergence. Our current trajectory designs explicitly avoid this issue. Future work will explore the use of nonlinear trajectories during ongoing-movement probing to more effectively eliminate this kinematic deadlock.

Third, the compensation coefficient $\eta$ was calibrated at a fixed grasping force of 30\,N using a continuous silicone-rubber fingertip. Its dependence on grasping force, material stiffness, and contact geometry has not been systematically characterized. A structural redesign of the fingertip offers a more fundamental path forward: replacing the bulk elastomer with an architected metamaterial, whose meso-scale topology can be engineered to produce distinct, repeatable deformation modes under compression, shear, and torsion, would encode multi-axial strain transmission directly into the mechanical domain, yielding more deterministic inertial signatures across varying operating conditions.

Finally, the two-phase trajectory planning strategy has been validated on a uniaxially-constrained mechanism. Extending this framework to tasks with more complex kinematic structures or highly nonlinear friction profiles remains an open direction for future research.

\section{Conclusion}
This paper presents \gls{tecdar}, a tactile sensing approach that localizes extrinsic contacts between a rigid grasped object—such as a plastic block, pen, book, or blackboard eraser—and its environment by capturing transient fingertip rotations with a miniature inertial sensor.
An event-triggered Bayes filter fuses these high-frequency gyroscopic signals with robot proprioception, achieving contact localization with errors ranging from 3.4\,mm to 12\,mm within approximately 180\,ms of contact onset, at a data throughput roughly two orders of magnitude below existing visuotactile methods. 
Moreover, building on this estimation capability, a two-phase trajectory planning strategy enables closed-loop manipulation of constrained mechanisms without prior kinematic knowledge, demonstrated on a physical paper cutter device.

More broadly, this work demonstrates that high-temporal-resolution inertial sensing can extract geometrically meaningful contact information from transient physical interactions that are too brief for conventional tactile sensing pipelines to resolve. We believe this capability, grounded in minimalist hardware and model-based estimation, offers a distinct and complementary dimension to tactile perception. By trading spatial resolution for temporal precision, it enables robots to react rapidly to unforeseen impacts within contact-rich and visually-occluded environments.

\bibliographystyle{IEEEtran}
\bibliography{ref}

@inproceedings{li2024IncipientSlipBasedRotation,
  title = {Incipient Slip-Based Rotation Measurement via Visuotactile Sensing During In-Hand Object Pivoting},
  booktitle = {2024 IEEE International Conference on Robotics and Automation (ICRA)},
  author = {Li, Mingxuan and Zhou, Yen Hang and Li, Tiemin and Jiang, Yao},
  date = {2024-05},
  year = {2024},
  pages = {17132--17138},
  url = {https://ieeexplore.ieee.org/document/10610988/},
  urldate = {2026-03-03},
  eventtitle = {2024 IEEE International Conference on Robotics and Automation (ICRA)},
  langid = {american}
}

@inproceedings{ma2021extrinsic,
  author    = {Ma, Daolin and Dong, Siyuan and Rodriguez, Alberto},
  title     = {Extrinsic Contact Sensing with Relative-Motion Tracking from Distributed Tactile Measurements},
  booktitle = {2021 IEEE International Conference on Robotics and Automation (ICRA)},
  year      = {2021},
  pages     = {9561781},
  doi       = {10.1109/ICRA48506.2021.9561781}
}

@inproceedings{kim2022ActiveExtrinsicContact,
  title={Active extrinsic contact sensing: Application to general peg-in-hole insertion},
  author={Kim, Sangwoon and Rodriguez, Alberto},
  booktitle={2022 International Conference on Robotics and Automation (ICRA)},
  pages={10241--10247},
  year={2022},
  organization={IEEE}
}

@inproceedings{higuera2023NeuralContactFields,
  author    = {Higuera, Carolina and Dong, Siyuan and Boots, Byron and Mukadam, Mustafa},
  title     = {Neural Contact Fields: Tracking Extrinsic Contact with Tactile Sensing},
  booktitle = {2023 IEEE International Conference on Robotics and Automation (ICRA)},
  year      = {2023},
  pages     = {12576--12582},
  doi       = {10.1109/ICRA48891.2023.10160526}
}

@inproceedings{sipos2022SimultaneousContactLocation,
  author    = {Sipos, Andrea and Fazeli, Nima},
  title     = {Simultaneous Contact Location and Object Pose Estimation Using Proprioception and Tactile Feedback},
  booktitle = {2022 IEEE/RSJ International Conference on Intelligent Robots and Systems (IROS)},
  year      = {2022},
  pages     = {3233--3240},
  doi       = {10.1109/IROS47612.2022.9981762}
}

@inproceedings{kim2023SimultaneousTactileEstimation,
  author    = {Kim, Sangwoon and Jha, Devesh K. and Romeres, Diego and Patre, Parag and Rodriguez, Alberto},
  title     = {Simultaneous Tactile Estimation and Control of Extrinsic Contact},
  booktitle = {2023 IEEE International Conference on Robotics and Automation (ICRA)},
  year      = {2023},
  pages     = {12563--12569}
}

@inproceedings{dong2021tactile,
  author    = {Dong, Siyuan and Jha, Devesh K. and Romeres, Diego and Kim, Sangwoon and Nikovski, Daniel and Rodriguez, Alberto},
  title     = {Tactile-RL for Insertion: Generalization to Objects of Unknown Geometry},
  booktitle = {2021 IEEE International Conference on Robotics and Automation (ICRA)},
  year      = {2021},
  pages     = {6437--6443}
}

@inproceedings{shiraiTactileToolManipulation2023,
  author    = {Shirai, Yuki and Jha, Devesh K. and Raghunathan, Arvind U. and Hong, Dennis},
  title     = {Tactile Tool Manipulation},
  booktitle = {2023 IEEE International Conference on Robotics and Automation (ICRA)},
  year      = {2023},
  pages     = {12597--12603}
}

@inproceedings{bronars2024TEXterityTactileExtrinsic,
  author    = {Bronars, Antonia and Kim, Sangwoon and Patre, Parag and Rodriguez, Alberto},
  title     = {TEXterity: Tactile Extrinsic deXterity},
  booktitle = {2024 IEEE International Conference on Robotics and Automation (ICRA)},
  year      = {2024},
  pages     = {7976--7983}
}

@article{zhaoTactileDrivenDexterousInHand2025,
  author    = {Zhao, Can and Xie, Lingzi and Huang, Bidan and Wang, Shuai and Ma, Daolin},
  title     = {Tactile-Driven Dexterous In-Hand Writing via Extrinsic Contact Sensing},
  journal   = {IEEE Robotics and Automation Letters},
  volume    = {10},
  number    = {9},
  pages     = {8914--8921},
  year      = {2025}
}

@online{wu2025TranTacLeveragingTransient,
  title = {TranTac: Leveraging Transient Tactile Signals for Contact-Rich Robotic Manipulation},
  shorttitle = {TranTac},
  author = {Wu, Yinghao and Hou, Shuhong and Zheng, Haowen and Li, Yichen and Lu, Weiyi and Zhou, Xun and Shao, Yitian},
  date = {2025-09-20},
  eprint = {2509.16550},
  eprinttype = {arXiv},
  eprintclass = {cs},
  url = {http://arxiv.org/abs/2509.16550},
  urldate = {2026-03-03},
  langid = {american},
  pubstate = {prepublished}
}

@article{bauza2023Tac2PoseTactileObjecta,
  title = {Tac2Pose: Tactile Object Pose Estimation from the First Touch},
  shorttitle = {Tac2Pose},
  author = {Bauza, Maria and Bronars, Antonia and Rodriguez, Alberto},
  year = {2023},
  month = nov,
  journal = {The International Journal of Robotics Research},
  volume = {42},
  number = {13},
  pages = {1185--1209},
  issn = {0278-3649, 1741-3176},
  urldate = {2025-08-19},
  langid = {english}
}

@inproceedings{ota2024TactileEstimationExtrinsic,
  title = {Tactile Estimation of Extrinsic Contact Patch for Stable Placement},
  booktitle = {2024 IEEE International Conference on Robotics and Automation (ICRA)},
  author = {Ota, Kei and Jha, Devesh K. and Jatavallabhula, Krishna Murthy and Kanezaki, Asako and Tenenbaum, Joshua B.},
  year = {2024},
  month = may,
  pages = {13876--13882},
  urldate = {2025-06-08},
  langid = {american}
}

@inproceedings{kimExtrinsicLineContact2025,
  title = {Extrinsic {{Line Contact Sensing}} from {{Visuo-Tactile Measurements}}},
  booktitle = {2025 22nd {{International Conference}} on {{Ubiquitous Robots}} ({{UR}})},
  author = {Kim, Yoonjin and Kim, Won Dong and Kim, Jung},
  date = {2025-06},
  pages = {46--50},
  urldate = {2025-08-20},
  eventtitle = {2025 22nd {{International Conference}} on {{Ubiquitous Robots}} ({{UR}})}
}

@online{leeViTaSCOPEVisuotactileImplicit2025,
  title = {{{ViTaSCOPE}}: {{Visuo-tactile Implicit Representation}} for {{In-hand Pose}} and {{Extrinsic Contact Estimation}}},
  shorttitle = {{{ViTaSCOPE}}},
  author = {Lee, Jayjun and Fazeli, Nima},
  date = {2025-06-13},
  eprint = {2506.12239},
  eprinttype = {arXiv},
  eprintclass = {cs},
  urldate = {2025-08-20},
  pubstate = {prepublished}
}

@inproceedings{ollerTactileDrivenNonPrehensileObject2024,
  title = {Tactile-Driven Non-Prehensile Object Manipulation via Extrinsic Contact Mode Control},
  booktitle={Robotics: Science and Systems},
  author = {Oller, Miquel and Berenson, Dmitry and Fazeli, Nima},
  date = {2024-07-15},
  year = {2024},
  urldate = {2025-08-20},
  eventtitle = {Robotics: {{Science}} and {{Systems}} 2024},
  isbn = {979-8-9902848-0-7},
  langid = {english}
}

@inproceedings{taylorObjectManipulationContact2023,
  title = {Object {{Manipulation Through Contact Configuration Regulation}}: {{Multiple}} and {{Intermittent Contacts}}},
  shorttitle = {Object {{Manipulation Through Contact Configuration Regulation}}},
  booktitle = {2023 {{IEEE}}/{{RSJ International Conference}} on {{Intelligent Robots}} and {{Systems}} ({{IROS}})},
  author = {Taylor, Orion and Doshi, Neel and Rodriguez, Alberto},
  date = {2023-10},
  year = {2024},
  pages = {8735--8743},
  issn = {2153-0866},
  urldate = {2025-08-20},
  eventtitle = {2023 {{IEEE}}/{{RSJ International Conference}} on {{Intelligent Robots}} and {{Systems}} ({{IROS}})},
  langid = {english}
}

@inproceedings{bicchi2000,
  author    = {Bicchi, Antonio and Kumar, Vijay},
  title     = {Robotic Grasping and Contact: A Review},
  booktitle = {Proceedings of the IEEE International Conference on Robotics and Automation (ICRA)},
  year      = {2000},
  pages     = {325--331}
}

@techreport{liang2024,
  title        = {Robust In-Hand Manipulation with Extrinsic Contacts},
  author       = {Liang, Boyuan and Ota, Kei and Tomizuka, Masayoshi and Jha, Devesh K.},
  year         = {2024},
  month        = {May},
  institution  = {Mitsubishi Electric Research Laboratories (MERL)},
  number       = {TR2024-061}
}

@book{bar2001estimation,
  title={Estimation with applications to tracking and navigation: theory algorithms and software},
  author={Bar-Shalom, Yaakov and Li, X Rong and Kirubarajan, Thiagalingam},
  year={2001},
  publisher={John Wiley \& Sons}
}

@article{van2026simultaneous,
  title={Simultaneous Extrinsic Contact and In-Hand Pose Estimation via Distributed Tactile Sensing},
  author={Van der Merwe, Mark and Ota, Kei and Berenson, Dmitry and Fazeli, Nima and Jha, Devesh K},
  journal={IEEE Robotics and Automation Letters},
  volume={11},
  number={3},
  pages={2394--2401},
  year={2026},
  publisher={IEEE}
}

@article{zhao2026tacman,
  title={Tacman-turbo: Proactive tactile control for robust and efficient articulated object manipulation},
  author={Zhao, Zihang and Qi, Zhenghao and Li, Yuyang and Cui, Leiyao and Han, Zhi and Ruan, Lecheng and Zhu, Yixin},
  journal={IEEE Transactions on Automation Science and Engineering},
  year={2026},
  publisher={IEEE}
}

@article{cravetz2025slip,
  author    = {Cravetz, Miranda and Vyas, Purva and Grimm, Casey and Davidson, Joseph R.},
  title     = {Slip detection for compliant robotic hands using inertial signals and deep learning},
  journal   = {Frontiers in Robotics and AI},
  volume    = {12},
  year      = {2025},
  doi       = {10.3389/frobt.2025.1698591}
}

@ARTICLE{funk2024evetac,
  author={Funk, Niklas and Helmut, Erik and Chalvatzaki, Georgia and Calandra, Roberto and Peters, Jan},
  journal={IEEE Transactions on Robotics}, 
  title={Evetac: An Event-Based Optical Tactile Sensor for Robotic Manipulation}, 
  year={2024},
  volume={40},
  number={},
  pages={3812-3832},
  doi={10.1109/TRO.2024.3428430}}

@article{lambeta2020digit,
  title={Digit: A novel design for a low-cost compact high-resolution tactile sensor with application to in-hand manipulation},
  author={Lambeta, Mike and Chou, Po-Wei and Tian, Stephen and Yang, Brian and Maloon, Benjamin and Most, Victoria Rose and Stroud, Dave and Santos, Raymond and Byagowi, Ahmad and Kammerer, Gregg and others},
  journal={IEEE Robotics and Automation Letters},
  volume={5},
  number={3},
  pages={3838--3845},
  year={2020},
  publisher={IEEE}
}

@article{cao2021six,
  title={Six-axis force/torque sensors for robotics applications: A review},
  author={Cao, Max Yiye and Laws, Stephen and y Baena, Ferdinando Rodriguez},
  journal={IEEE Sensors Journal},
  volume={21},
  number={24},
  pages={27238--27251},
  year={2021},
  publisher={IEEE}
}

@article{luo2021learning,
  title={Learning human--environment interactions using conformal tactile textiles},
  author={Luo, Yiyue and Li, Yunzhu and Sharma, Pratyusha and Shou, Wan and Wu, Kui and Foshey, Michael and Li, Beichen and Palacios, Tom{\'a}s and Torralba, Antonio and Matusik, Wojciech},
  journal={Nature Electronics},
  volume={4},
  number={3},
  pages={193--201},
  year={2021},
  publisher={Nature Publishing Group UK London}
}

@article{yuan2017gelsight,
  title={Gelsight: High-resolution robot tactile sensors for estimating geometry and force},
  author={Yuan, Wenzhen and Dong, Siyuan and Adelson, Edward H},
  journal={Sensors},
  volume={17},
  number={12},
  pages={2762},
  year={2017},
  publisher={MDPI}
}

@online{yajima2026Touch2InsertZeroShotPeg,
  title = {{{Touch2Insert}}: {{Zero-Shot Peg Insertion}} by {{Touching Intersections}} of {{Peg}} and {{Hole}}},
  shorttitle = {{{Touch2Insert}}},
  author = {Yajima, Masaru and Shin, Yuma and Kawakami, Rei and Kanezaki, Asako and Ota, Kei},
  date = {2026-03-04},
  eprint = {2603.03627},
  eprinttype = {arXiv},
  eprintclass = {cs},
  urldate = {2026-03-08},
  langid = {american},
  pubstate = {prepublished}
}

@inproceedings{taunyazov2021extended,
  title={Extended tactile perception: Vibration sensing through tools and grasped objects},
  author={Taunyazov, Tasbolat and Song, Luar Shui and Lim, Eugene and See, Hian Hian and Lee, David and Tee, Benjamin CK and Soh, Harold},
  booktitle={2021 IEEE/RSJ International Conference on Intelligent Robots and Systems (IROS)},
  pages={1755--1762},
  year={2021},
  organization={IEEE}
}

@incollection{schurmann2012high,
  title={A high-speed tactile sensor for slip detection},
  author={Sch{\"u}rmann, Carsten and Sch{\"o}pfer, Matthias and Haschke, Robert and Ritter, Helge},
  booktitle={Towards Service Robots for Everyday Environments: Recent Advances in Designing Service Robots for Complex Tasks in Everyday Environments},
  pages={403--415},
  year={2012},
  publisher={Springer}
}

@inproceedings{bhirangi2025anyskin,
  title={Anyskin: Plug-and-play skin sensing for robotic touch},
  author={Bhirangi, Raunaq and Pattabiraman, Venkatesh and Erciyes, Enes and Cao, Yifeng and Hellebrekers, Tess and Pinto, Lerrel},
  booktitle={2025 IEEE International Conference on Robotics and Automation (ICRA)},
  pages={16563--16570},
  year={2025},
  organization={IEEE}
}

@article{posa2014direct,
  title={A direct method for trajectory optimization of rigid bodies through contact},
  author={Posa, Michael and Cantu, Cecilia and Tedrake, Russ},
  journal={The International Journal of Robotics Research},
  volume={33},
  number={1},
  pages={69--81},
  year={2014},
  publisher={Sage Publications Sage UK: London, England}
}

@inproceedings{manchester2019variational,
  title={Variational contact-implicit trajectory optimization},
  author={Manchester, Zachary and Kuindersma, Scott},
  booktitle={Robotics Research: The 18th International Symposium ISRR},
  pages={985--1000},
  year={2019},
  organization={Springer}
}

@article{chavan2020planar,
  title={Planar in-hand manipulation via motion cones},
  author={Chavan-Dafle, Nikhil and Holladay, Rachel and Rodriguez, Alberto},
  journal={The International Journal of Robotics Research},
  volume={39},
  number={2-3},
  pages={163--182},
  year={2020},
  publisher={SAGE Publications Sage UK: London, England}
}

@inproceedings{suh2022seed,
  title={Seed: Series elastic end effectors in 6d for visuotactile tool use},
  author={Suh, HJ Terry and Kuppuswamy, Naveen and Pang, Tao and Mitiguy, Paul and Alspach, Alex and Tedrake, Russ},
  booktitle={2022 IEEE/RSJ International Conference on Intelligent Robots and Systems (IROS)},
  pages={4684--4691},
  year={2022},
  organization={IEEE}
}

@article{huo2026recent,
  title={Recent advances of flexible pressure tactile sensors: Sensing mechanisms, performance breakthroughs, and intelligent applications},
  author={Huo, Xiaoqing and Liu, Bin and Wu, Zhiyi},
  journal={Advanced Materials Technologies},
  volume={11},
  number={5},
  pages={e01837},
  year={2026},
  publisher={Wiley Online Library}
}

@article{yin2025gelevent,
  title={GelEvent—A novel high-speed tactile sensor with event camera},
  author={Yin, Dong and Lu, Siliang and Yang, Jun and Zhang, Yang and Dai, Zhiwei and Nan, De and Cai, Bolin and He, Shuping and Chen, Xiangcheng},
  journal={IEEE Transactions on Instrumentation and Measurement},
  year={2025},
  publisher={IEEE}
}

\end{document}